\documentclass[final,12pt,authoryear]{elsarticle}

\usepackage{graphicx}
\usepackage{xcolor}

\usepackage{amsmath}
\usepackage{amsthm}
\usepackage{amssymb}
\usepackage{amsfonts}
\usepackage{optidef}

\theoremstyle{definition}
\newtheorem{definition}{Definition}
\theoremstyle{remark}
\newtheorem{remark}{Remark}

\usepackage{algorithmic}
\usepackage{array}
\usepackage{booktabs}
\usepackage{multirow}
\usepackage{comment}
\usepackage{makecell}
\usepackage{enumitem}

\usepackage{subcaption}

\usepackage{stfloats}
\usepackage{bm}
\usepackage{float}

\usepackage{url}
\usepackage[hidelinks]{hyperref}

\newcommand{\tablefontsize}{\small}
\newcommand{\tablecolsep}{4pt}

\newlength{\singlefigwidth}
\newlength{\halffigwidth}
\newlength{\thirdfigwidth}
\newlength{\fullfigwidth}
\DeclareMathOperator{\RG}{RG}
\newcommand{\model}{URC$^2$}

\makeatletter
\def\ps@pprintTitle{%
  \let\@oddhead\@empty
  \let\@evenhead\@empty
  \let\@oddfoot\@empty
  \let\@evenfoot\@empty}
\makeatother

\begin{document}

\begin{frontmatter}

\title{Developing a Totally Unimodular Linear Program for Optimal Conformance Checking: When and Why It Complements \texorpdfstring{$A^*$}{A-star}}

\author[biu]{Izack Cohen}
\ead{izack.cohen@biu.ac.il}

\affiliation[biu]{
  organization={Faculty of Engineering, Bar-Ilan University},
  city={Ramat Gan},
  country={Israel}
}

\begin{abstract}
Alignment-based conformance checking is the state-of-the-art approach for comparing observed process executions with normative process models. The standard exact solution relies on an $A^*$-based heuristic search, which can exhibit exponential runtime in the presence of long traces or substantial deviations.
This paper introduces a reformulation of alignment-based conformance checking as a totally unimodular linear program (LP) defined on the reachability graph of the synchronous product. By exploiting the underlying network-flow structure, the proposed formulation guarantees the existence of an integral optimal extreme-point solution through LP relaxation, thereby avoiding the combinatorial overhead associated with integer variables and branch-and-bound search.

We conduct an extensive empirical evaluation on more than 2.1 million conformance checking instances derived from real-world and synthetic benchmark datasets. The results show that $A^*$
 and the LP approach exhibit complementary performance characteristics: the former performs best on short, well-conforming traces, while the LP formulation provides substantial speedups for longer traces with deviations, precisely where conformance checking is most informative. Based on these findings, we derive simple algorithm-selection guidelines that combine both approaches, achieving average runtime savings of 38.6\% with 96\% selection accuracy compared to always using $A^*$.
\end{abstract}


\end{frontmatter}
{\footnotesize
\noindent\textbf{Author-accepted manuscript.}
Accepted for publication in \emph{Expert Systems with Applications}.
This version is available under CC BY-NC-ND 4.0.
Please cite the published article.
\par}

\vspace{0.75em}

\vspace{0.75em}
\section{Introduction}\label{sec:intro}

Conformance checking is a fundamental task in process mining that quantifies how well observed process executions align with a normative process model. Among the available techniques, alignment-based conformance checking has become the standard due to its semantic rigor and diagnostic power, as it computes an optimal correspondence between an observed trace and a process model. Computing such optimal alignments, however, is computationally demanding, and practical solutions typically rely on heuristic search algorithms such as $A^*$ or on mixed-integer linear programming (MILP) formulations.

A central question in this domain is whether exact conformance checking necessarily requires specialized search algorithms and custom implementations, or whether it can be expressed using standard off-the-shelf optimization paradigms that benefit from decades of advances in operations research. While approaches such as $A^*$ can be highly effective in favorable cases, they require careful heuristic design, custom implementations, and may exhibit exponential behavior on long or highly deviating traces. In contrast, problems that admit a formulation within established optimization frameworks can leverage mature solver technology, allowing direct use of off-the-shelf solvers without custom implementation effort, provided that their underlying structure supports \textit{efficient} exact solution methods.

This paper shows that alignment-based conformance checking admits such a formulation. We present a reformulation of the problem as a linear program defined on the reachability graph of the synchronous product, whose constraint matrix is totally unimodular. We refer to this formulation as the Unimodular Reformulation for Conformance Checking (\model). The key theoretical insight is that the resulting problem corresponds to a minimum-cost network flow with integral right-hand sides, which guarantees that every extreme point of the LP feasible region is integral. In other words, whenever an optimum exists, the LP admits an integral optimal solution. As a consequence, optimal alignments can be computed using standard LP solvers, without integer variables or branch-and-bound search.

Importantly, this result does not eliminate the inherent state-space complexity of conformance checking: the reachability graph of the synchronous product may still grow exponentially in the worst case. Rather, \model{} removes combinatorial optimization from within the explored state space by replacing heuristic or integer search with a single linear optimization problem. In this sense, \model{} shifts the computational burden from search control to linear optimization over an explicitly constructed portion of the state space, while relying on solver efficiency rather than heuristic exploration.

In practice, the current \model{} implementation constructs an explicitly bounded
subgraph of the reachability graph of the synchronous product by breadth-first
exploration up to a practical depth limit. This bound is used to control graph construction cost and avoid unbounded expansion in the presence of cycles or
concurrency. The LP is then solved exactly on the constructed subgraph.

Rather than proposing \model{} as a universal replacement for existing methods, we position it as a complement to $A^*$. Through an extensive empirical study comprising more than 2.1 million conformance-checking instances across five real-world and synthetic benchmark datasets, we show that neither approach dominates. $A^*$ performs best on short, well-conforming traces, where heuristic guidance keeps the explored state space small. In contrast, \model{} provides substantial performance gains on longer traces with deviations, precisely where heuristic search tends to degrade. This distinction is practically relevant since conformance checking is most informative in scenarios where deviations are expected, while perfectly conforming executions require little analysis.

Based on these observations, we identify the conditions under which each approach is preferable and derive a simple algorithm-selection rule that combines both methods. This hybrid strategy yields 38.6\% runtime savings compared to consistently applying $A^*$ alone, while preserving optimality.

\vspace{10pt}
The main contributions of this paper are:
\begin{enumerate}
\item \textit{A totally unimodular LP formulation.} We reformulate alignment-based conformance checking as a minimum-cost network flow problem on the reachability graph of the synchronous product and show that its constraint matrix is totally unimodular. This guarantees the existence of an integral optimal extreme-point solution via linear programming, enabling exact alignment computation without integer variables.

\item \textit{Large-scale empirical evaluation.} We conduct an extensive experimental study comparing \model{} and $A^*$ across more than 2.1 million instances from real-world and synthetic datasets, analyzing performance across trace lengths, model characteristics, and deviation levels.

\item \textit{Practical algorithm-selection guidelines.} We identify the key factors governing the relative performance of $A^*$ and \model{}, and provide a simple, actionable selection rule that achieves 96\% selection accuracy. The results demonstrate that the two approaches are complementary: \model{} achieves substantial speedups on long, non-conforming traces, while $A^*$ remains preferable for short, well-conforming ones.

\end{enumerate}

The remainder of this paper is structured as follows. Section~\ref{sec:related} reviews related work. Section~\ref{sec:concepts} introduces the required concepts and notation. Section~\ref{sec:model} presents the \model{} formulation and establishes its total unimodularity, followed by an illustrative example in Section~\ref{sec:example-mcf}. Section~\ref{sec:experiments} describes the experimental design, and Section~\ref{sec:results} reports the empirical results. Based on these findings, Section~\ref{sec:algorithm-selection} provides practical guidelines for choosing between $A^*$ and \model{}. Section~\ref{sec:limitations} discusses limitations and future research directions, and Section~\ref{sec:conclusion} concludes the paper.

\section{Related Work}\label{sec:related}

Since its formalization, conformance checking has been studied through exact, approximate, decomposition-based, and symbolic approaches. This section reviews the most relevant research streams and positions the proposed unimodular LP formulation within this body of work.

\subsection{Exact Alignment-Based Conformance Checking}

The notion of alignments was introduced by \citet{adriansyah2011conformance} and later formalized in \citet{Adriansyah2014}, where optimal alignments between traces and Petri net models are computed using $A^*$ search on the synchronous product. This formulation has become the standard approach for exact conformance checking and is supported by mature implementations in process mining frameworks such as \texttt{ProM} and \texttt{PM4Py}. The approach relies on heuristic functions derived from the marking equation of the underlying Petri net. Subsequent work by \citet{vandongen2018efficiently} introduced an extended marking equation with tighter bounds, reducing the number of linear programs solved during search, particularly for traces with swapped activities. \citet{li2022cache} extended this with a caching strategy for split-point-based alignment calculation that improves overall search efficiency. More recently, \citet{casas2024reach} proposed REACH, which reduces both the number of explored states and the processing time per state through mandatory-move forcing and partial reachability graph caching of the process model. Despite these improvements, $A^*$-based methods preserve the fundamental architecture of best-first search and may exhibit exponential complexity in the branching factor and alignment length, which can make them impractical for traces with many deviations or models with high concurrency.

In parallel, mathematical programming approaches have been explored for conformance checking. \citet{de2013aligning} proposed a multi-perspective approach based on mixed-integer linear programming (MILP) to incorporate control-flow, data, resources, and time. More recently, \citet{schwanen2024process} formulated the alignment problem for block-structured process trees as a MILP, showing that the problem lies in $\mathsf{NP}$ for this restricted model class. Their formulation requires integer variables to encode parallelism and reduces to a polynomial-time linear program only when parallel operators are absent.

Our approach differs in scope and mechanism. Rather than restricting the model class, \model{} supports general workflow Petri nets and formulates alignment computation as a minimum-cost network flow problem defined on the reachability graph of the synchronous product. The key insight is that the resulting constraint matrix is totally unimodular, which guarantees the existence of an integral optimal extreme-point solution through linear programming alone, without introducing integer variables, independently of the degree of concurrency. Consequently, the resulting formulation admits polynomial-time solvability in the size of the reachability graph.

\subsection{Approximate and Heuristic Methods}

Approximation and heuristic techniques trade accuracy for scalability. 
\citet{schuster2020alignment} proposed alignment approximation for process trees by 
exploiting their hierarchical structure. Sampling-based methods~\citep{bauer2019estimating,
fani2020conformance} provide statistical guarantees on aggregate conformance measures but sacrifice trace-level diagnostics.
Heuristic approaches, including evolutionary algorithms~\citep{buijs2014flexible,
taymouri2018evolutionary}, achieve scalability at the cost of optimality guarantees. 
Data-structure optimizations such as trie-based caching~\citep{9576845} and 
tandem-repeat compression~\citep{reissner2020scalable} reduce redundant computation but depend on trace similarity or repetitive patterns.

\subsection{Decomposition Strategies}

Model-based decomposition partitions process models into fragments that are aligned 
independently~\citep{munoz2014single,taymouri2016recursive,cheng2023optimal}. Log-based decomposition 
partitions event logs horizontally or vertically for distributed 
processing~\citep{valencia2021empowering, bogdanov2024scalable}. These approaches trade global optimality 
for tractability and may introduce recomposition overhead.

\subsection{Symbolic Encodings}

Symbolic techniques compact the state space using decision diagrams~\citep{bloemen2018} 
or SAT/MaxSAT formulations~\citep{boltenhagen2021optimized}. While effective for pruning, they may incur high memory consumption on complex models.

\subsection{Positioning}

The unimodular LP formulation \model{} occupies a distinct position within alignment-based conformance checking research. Unlike decomposition or approximation approaches, \model{} finds optimal alignments. Unlike $A^*$-based methods, which may exhibit exponential behavior on nonconforming traces, the LP formulation has polynomial complexity in the size of the constructed reachability graph. By exploiting the structural properties of the constraint matrix, \model{} avoids the combinatorial complexity of MILP formulations and enables direct use of standard off-the-shelf LP solvers, without branch-and-bound or cutting-plane procedures.

At the same time, \model{} preserves the diagnostic power of alignment-based conformance checking, in contrast to token-based or sampling approaches. Its main computational overhead lies in constructing a bounded portion of the reachability graph on which the LP is solved. Our empirical results reveal complementary performance behavior: $A^*$ excels on short, well-conforming traces, whereas \model{} provides substantial advantages on longer traces with deviations. This complementarity has practical implications, as conformance checking is most informative precisely when deviations occur, and motivates algorithm-selection guidelines that exploit the strengths of both approaches.

\section{Concepts and Notation}\label{sec:concepts}
This section introduces concepts required for developing \model, starting with defining process and trace models, used for representing expected and observed process behaviors, respectively. Throughout this section, we use standard process mining definitions and notation \citep[see e.g.,][]{van2016data}, with some adaptations. Next, we explain how to merge a process model and a trace as a synchronous product followed by defining a cost function and key alignment concepts.

\begin{definition}[Labeled Marked Petri Net]
A labeled marked Petri net is a tuple
\[
N = (P, T, F, \lambda, m_i, m_f),
\]
where:
\begin{description}
  \item[$P$:] finite set of places,
  \item[$T$:] finite set of transitions with $P \cap T = \emptyset$,
  \item[$F$:] set of directed arcs (flow relations), with 
  $F \subseteq (P \times T) \cup (T \times P)$,
  \item[$\lambda$:] labeling function $\lambda: T \to A \cup \{\tau\}$ assigning each transition
  either an activity label or the silent label $\tau \notin A$, where $A$ is the set of activities,
  \item[$m_i$:] initial marking (initial token distribution),
  \item[$m_f$:] final marking (completion token distribution).
\end{description}
\end{definition}

We use the above definition to represent a process model, where places correspond to conditions or states, and transitions to events or activities. The set of flow relations \(F\) defines how tokens can move between places and transitions. The labeling function \(\lambda\) maps each transition to an activity; the special label \(\tau\) indicates a silent (unobservable or internal) transition.

\begin{figure*}[htbp]
\centering
\begin{subfigure}[b]{0.42\textwidth}
    \centering
    \includegraphics[width=\textwidth]{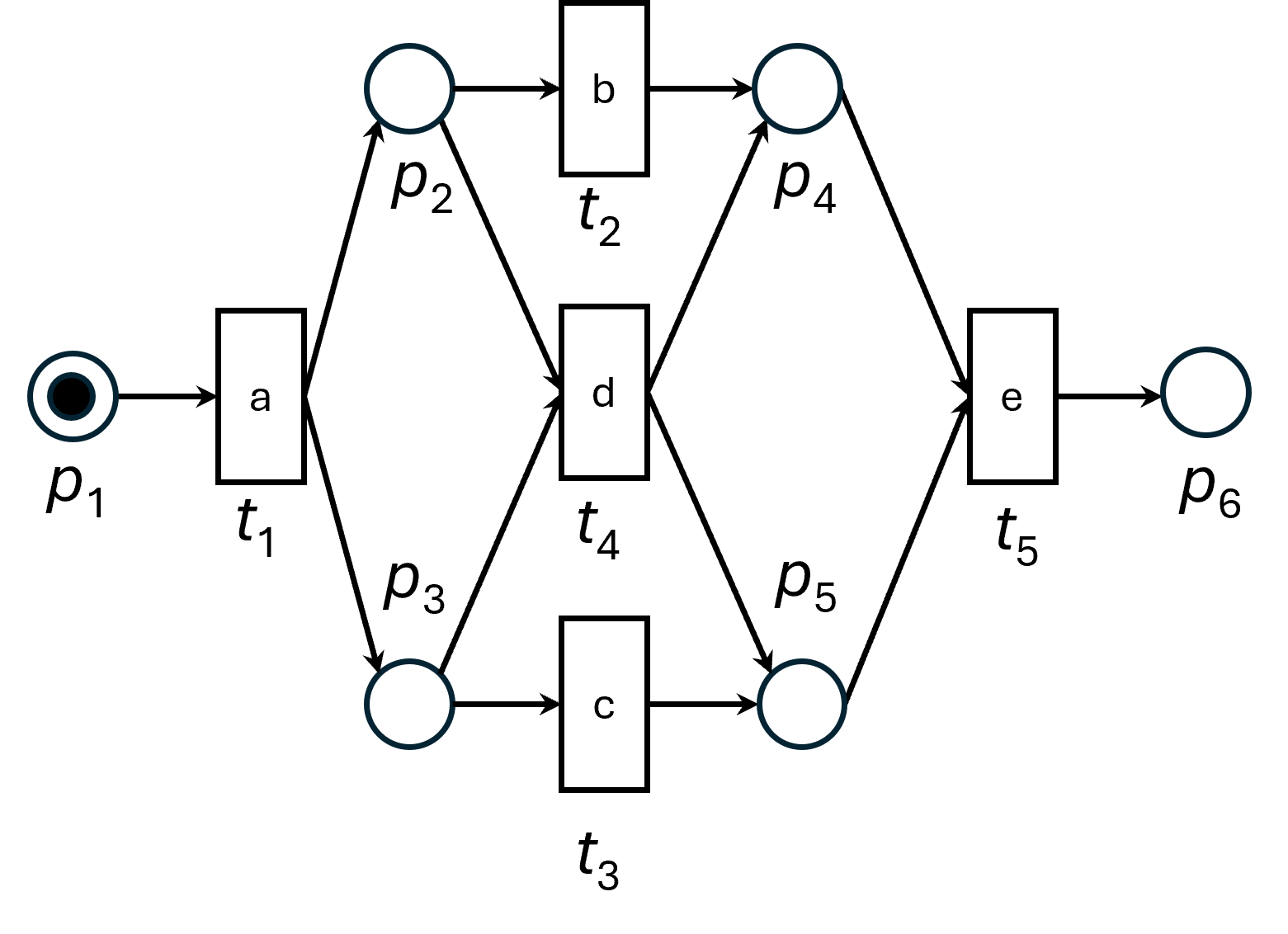}
    \caption{Acyclic process model}
    \label{fig_acyclic}
\end{subfigure}
\hfill
\begin{subfigure}[b]{0.42\textwidth}
    \centering
    \includegraphics[width=\textwidth]{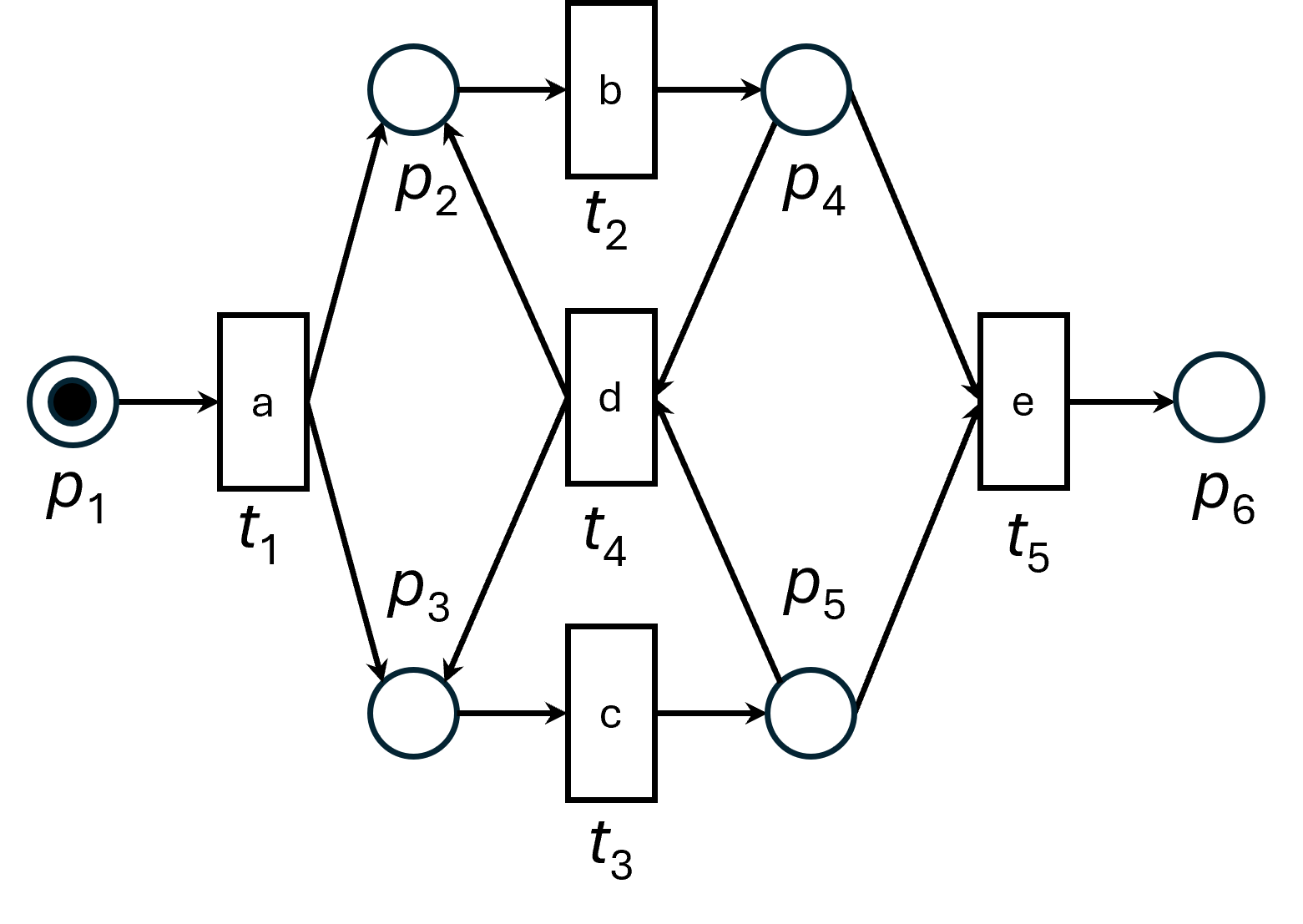}
    \caption{Cyclic process model}
    \label{fig_cyclic}
\end{subfigure}
\caption{Two Petri nets, without and with a loop.}
\label{fig_toy}
\end{figure*}

\begin{figure*}[htbp]
\centering
\begin{subfigure}[b]{0.75\textwidth}
    \centering
    \includegraphics[width=\textwidth]{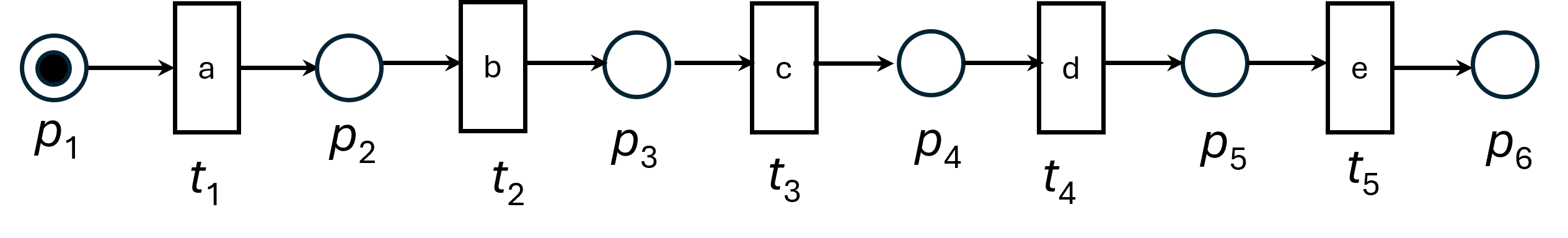}
    \caption{}
    \label{fig_toy_trace1}
\end{subfigure}\\[0.8em]
\begin{subfigure}[b]{0.42\textwidth}
    \centering
    \includegraphics[width=\textwidth]{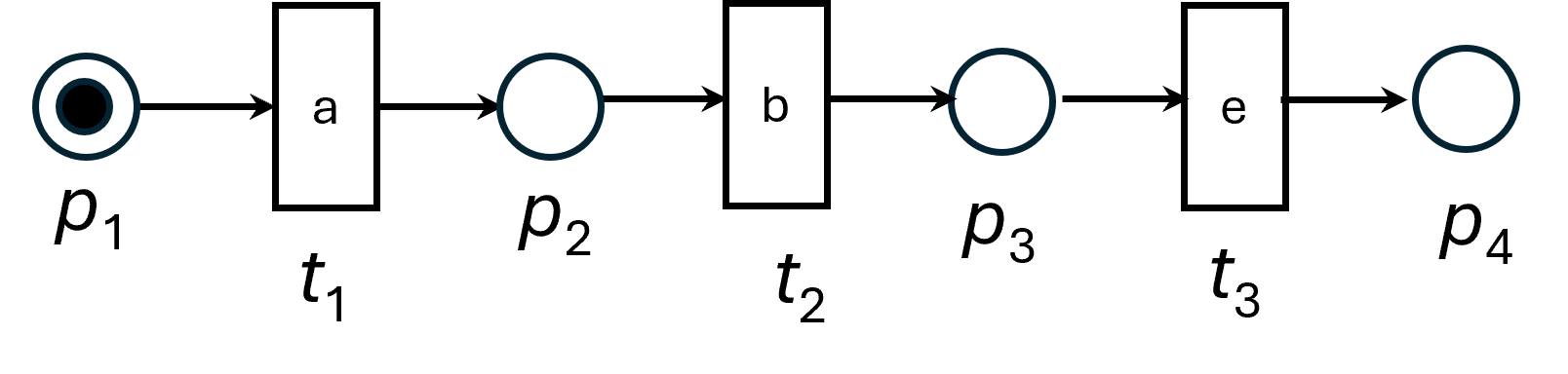}
    \caption{}
    \label{fig_toy_trace2}
\end{subfigure}\\[0.8em]
\begin{subfigure}[b]{0.85\textwidth}
    \centering
    \includegraphics[width=\textwidth]{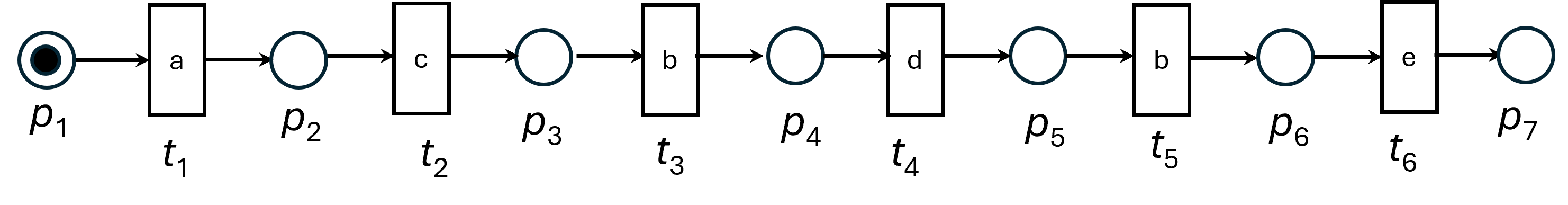}
    \caption{}
    \label{fig_toy_trace3}
\end{subfigure}
\caption{Three models for traces (a) $\langle abcde \rangle$ (b) $\langle abe \rangle$ (c) $\langle acbdbe \rangle$.}
\label{fig_toytrace}
\end{figure*}

Figure~\ref{fig_toy} presents two process models. The one in Figure~\ref{fig_acyclic} does not include loops and the other, in Figure~\ref{fig_cyclic}, does. Mathematically, for both processes $P=\{p_1,p_2,\dots,p_6 \}$, $T=\{t_1,t_2,\dots,t_5 \}$, $m_i=[1,0,0,0,0,0]$,
$m_f=[0,0,0,0,0,1]$, $A=\{a,b,c,d,e \}$ and $\lambda(t_1)=a, \lambda(t_2)=b, \lambda(t_3)=c,\lambda(t_4)=d,\lambda(t_5)=e$. The sets of flows are different: 

\[
F_{\ref{fig_acyclic}} =
\left\{
\begin{aligned}
  & (p_1,t_1), (t_1,p_2), (t_1,p_3), (p_2,t_2), \\
  & (p_2,t_4), (p_3,t_3), (p_3,t_4), (t_2,p_4), \\
  & (t_3,p_5), (t_4,p_4), (t_4,p_5), (p_4,t_5), \\
  & (p_5,t_5), (t_5,p_6)
\end{aligned}
\right\}, \text{and}
\]
\[
F_{\ref{fig_cyclic}} =
\left\{
\begin{aligned}
  & (p_1,t_1), (t_1,p_2), (t_1,p_3), (p_2,t_2), \\
  & (t_4,p_2), (p_3,t_3), (t_4,p_3), (t_2,p_4), \\
  & (t_3,p_5), (p_4,t_4), (p_5,t_4), (p_4,t_5), \\
  & (p_5,t_5), (t_5,p_6)
\end{aligned}
\right\} \text{.}
\]

Table~\ref{tab:incidence-toy} presents the models' \textit{incidence matrices}, where rows correspond to places and columns to transitions. Each entry $I_{ij}$ indicates the net change in tokens at place $p_i$ when transition $t_j$ fires with $-1$ denoting token consumption, $+1$ token production, and $0$ indicating no effect. For example, $I_{11}=-1$ corresponds to a token flowing from $p_1$ and consumed in $t_1$, interpreted as flow $(p_1,t_1)$. $I_{21}=1$ represents a token produced by $t_1$ and moving into $p_2$, related to flow $(t_1,p_2)$. 

\begin{table}[htbp]
\centering
\small
\setlength{\tabcolsep}{5pt}
\caption{Incidence matrices for the process models in Figure~\ref{fig_toy}.}
\label{tab:incidence-toy}
\begin{tabular}{c|rrrrr}
\multicolumn{6}{c}{(\ref{fig_acyclic}) Acyclic} \\[2mm]
\toprule
& $t_1$ & $t_2$ & $t_3$ & $t_4$ & $t_5$ \\
\midrule
$p_1$ & $-1$ &  0 &  0 &  0 &  0 \\
$p_2$ &  1 & $-1$ &  0 & $-1$ &  0 \\
$p_3$ &  1 &  0 & $-1$ & $-1$ &  0 \\
$p_4$ &  0 &  1 &  0 &  1 & $-1$ \\
$p_5$ &  0 &  0 &  1 &  1 & $-1$ \\
$p_6$ &  0 &  0 &  0 &  0 &  1 \\
\bottomrule
\end{tabular}
\hspace{1.5cm}
\begin{tabular}{c|rrrrr}
\multicolumn{6}{c}{(\ref{fig_cyclic}) Cyclic} \\[2mm]
\toprule
& $t_1$ & $t_2$ & $t_3$ & $t_4$ & $t_5$ \\
\midrule
$p_1$ & $-1$ &  0 &  0 &  0 &  0 \\
$p_2$ &  1 & $-1$ &  0 &  1 &  0 \\
$p_3$ &  1 &  0 & $-1$ &  1 &  0 \\
$p_4$ &  0 &  1 &  0 & $-1$ & $-1$ \\
$p_5$ &  0 &  0 &  1 & $-1$ & $-1$ \\
$p_6$ &  0 &  0 &  0 &  0 &  1 \\
\bottomrule
\end{tabular}
\end{table}

\newpage
\begin{definition}[Trace Model]
Let $A$ be a finite set of activities. Let $A'$ denote the set of all finite sequences over $A$, representing possibly completed traces. Assume that $\sigma \in A'$ is a trace of length $n$, then its corresponding trace model is defined as the Petri net
\[
TN = (P, T, F, \lambda, m_i, m_f),
\]
where $P = \{p_0, p_1, \dots, p_n\}$ is the set of places, 
$T = \{t_1, t_2, \dots, t_n\}$ is the set of transitions, and
\[
F = \{(p_i, t_{i+1}) \mid 0 \le i < n\} \;\cup\; \{(t_i, p_i) \mid 1 \le i \le n\}
\]
is the set of flow relations. The initial and final markings are given by $m_i = [p_0]$ and $m_f = [p_n]$, respectively. 
The labeling function $\lambda : T \to A$ assigns each transition its corresponding activity label, i.e., $\lambda(t_i) = \sigma(i)$ for $1 \le i \le n$.
\end{definition}

A trace model represents the sequence of events observed during an execution, with its structure derived directly from the trace 
$\sigma$. Each transition corresponds to an event, while the initial and final markings define the start and end of the execution. This structure enables direct comparison with the expected process behavior when integrated into a synchronous product.
Figure~\ref{fig_toytrace} illustrates this by presenting three traces, (a) $\langle abcde \rangle$ (b) $\langle abe \rangle$ (c) $\langle acbdbe \rangle$, as trace models.

\begin{definition}[Synchronous Product]\label{def:synch_prod}
Consider a process model
\[
SN = (P^{SN}, T^{SN}, F^{SN}, \lambda^{SN}, m_i^{SN}, m_f^{SN}),
\]
and a trace model
\[
TN = (P^{TN}, T^{TN}, F^{TN}, \lambda^{TN}, m_i^{TN}, m_f^{TN})
\]
for an observed trace $\sigma$. The synchronous product is defined as
\[
SP = (P, T, F, \lambda, m_i, m_f),
\]
where
\begin{description}[itemsep=0.1em, topsep=0pt]

\item[$P$:]
$P = P^{SN} \cup P^{TN}$.

\item[$T$:]
The transition set is partitioned as
\[
T = T^{MM} \cup T^{LM} \cup T^{SM},
\]
where
\[
\begin{aligned}
T^{MM} &= \{(t,\gg) \mid t \in T^{SN}\},\\
T^{LM} &= \{(\gg,t) \mid t \in T^{TN}\},\\
T^{SM} &= \{(t_1,t_2) \in T^{SN} \times T^{TN} \mid 
\lambda^{SN}(t_1) = \lambda^{TN}(t_2)\}.
\end{aligned}
\]

\item[$F$:]
The flow relation is defined as
\[
\begin{aligned}
F =\;& \{(p,(t_1,t_2)) \mid (p,t_1)\in F^{SN} \ \text{or}\ (p,t_2)\in F^{TN}\} \\
&\cup \{((t_1,t_2),p) \mid (t_1,p)\in F^{SN} \ \text{or}\ (t_2,p)\in F^{TN}\}.
\end{aligned}
\]

\item[$m_i, m_f$:]
The initial and final markings are given by
\[
m_i = m_i^{SN} + m_i^{TN},
\qquad
m_f = m_f^{SN} + m_f^{TN}.
\]

\item[$\lambda$:]
The labeling function is defined by
\[
\lambda((t_1,t_2)) = (\ell_1,\ell_2),
\]
where
\[
\ell_1 =
\begin{cases}
\lambda^{SN}(t_1), & \text{if } t_1 \in T^{SN},\\
\gg, & \text{otherwise},
\end{cases}
\qquad
\ell_2 =
\begin{cases}
\lambda^{TN}(t_2), & \text{if } t_2 \in T^{TN},\\
\gg, & \text{otherwise}.
\end{cases}
\]

\end{description}
\end{definition}
The synchronous product is a Petri net that integrates the process and trace models, allowing for a detailed comparison of their behaviors. By categorizing transitions into synchronous moves (where the process and trace align), model moves (indicating missing trace events), and log moves (representing unexpected trace events), this approach provides a structured foundation for conformance checking. To quantify deviations, a cost function assigns a numerical penalty to each type of move, capturing its impact on alignment quality.

\newpage
\begin{definition}[Cost Function]\label{def:cost_func}
For the synchronous product\\ \(SP = (P, T, F, \lambda, m_i, m_f)\), the \emph{cost function} \(c: T \to \mathbb{R}_{\ge0}\) assigns a non-negative cost to each transition as follows:
\begin{description}[itemsep=0.1em, topsep=0pt]
    \item For every synchronous move \((t_1,t_2) \in T^{SM}\), set \(c((t_1,t_2)) = 0\).
    \item For a model move \((t,\gg) \in T^{MM}\) with \(\lambda^{SN}(t) = \tau\), set \(c((t,\gg)) = \epsilon\), where \(\epsilon \to 0^+\).
    \item For all other moves (i.e., nonsynchronous model or log moves), assign a cost of 1.
\end{description}
\end{definition}

The cost function measures deviations between modeled and observed behavior, assigning no penalty to synchronous moves while penalizing discrepancies. It plays a key role in determining the optimal alignment between the trace and the process model.

\begin{definition}[Optimal Alignment]\label{def:opt_alignment}
Consider $A$ a set of activities, $\sigma \in A'$ a trace with a trace model $TN$, $SN$ as a process model, and $SP$ as the synchronous product of $SN$ and $TN$. Let $ c : T \rightarrow \mathbb{R}_{\ge0}$ be a cost function. Then, an optimal alignment $\gamma^{opt} \in L_{SP}$, where $L_{SP}$ is the set of possible execution sequences, is a full execution sequence of $SP$ such that for all $\gamma \in L_{SP}$, $c(\gamma^{opt}) \leq c(\gamma)$, where $c(\gamma) = \sum_{1\leq i \leq  |\gamma|} \, c(\gamma(i))$.
\end{definition}
In other words, an optimal alignment is the sequence of moves through the synchronous product that minimizes the total cost, representing the best possible correspondence between the observed trace and the expected process behavior.

\begin{figure}[ht]
\centering
\includegraphics[width=0.6\textwidth]{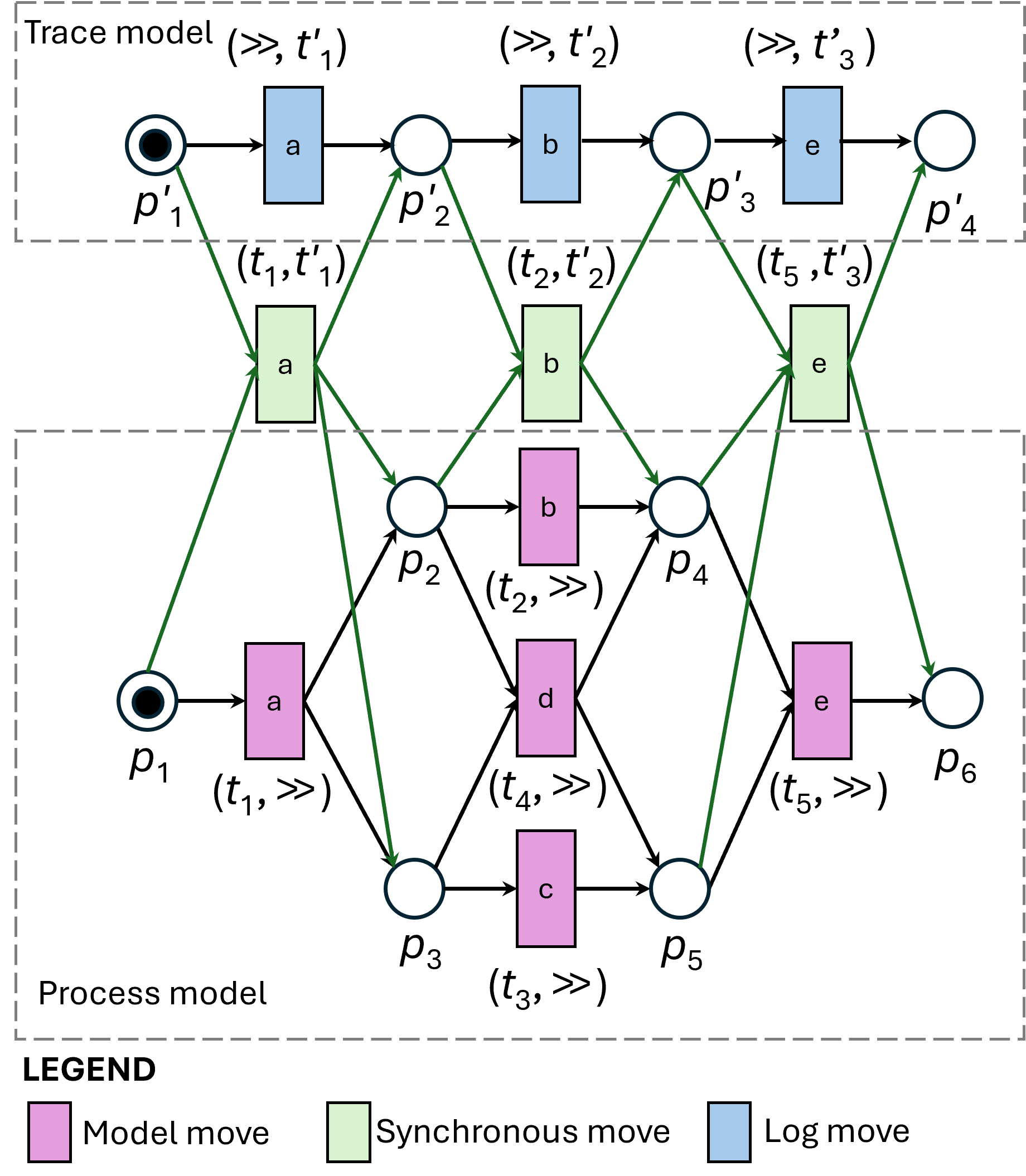}
\caption{A synchronous product of process and trace models.}
\label{fig_synch_prod}
\end{figure}

To make these concepts concrete, we illustrate them using the process model in Figure~\ref{fig_acyclic} and the trace in Figure~\ref{fig_toy_trace2}. 
Applying Definition~\ref{def:synch_prod} yields the synchronous product shown in Figure~\ref{fig_synch_prod}. 
A standard cost function, by Definition~\ref{def:cost_func}, assigns a unit cost to each log or model move, and zero cost to synchronous moves. 
Consequently, finding an optimal alignment according to Definition~\ref{def:opt_alignment} amounts to transferring the tokens from the initial state of the synchronous product, $[p_1,p'_1]$, to its final state, $[p_6,p'_4]$, while minimizing the number of non-synchronous moves. 
The two optimal alignments are,
\begin{align}
\gamma^{\mathrm{opt}} = \{&
\langle (t_1,t'_1), (t_2,t'_2), (t_3,\gg), (t_5,t'_3) \rangle, \notag\\
&\langle (t_1,t'_1), (t_3,\gg), (t_2,t'_2), (t_5,t'_3) \rangle \}.
\label{eq:opt_alignment}
\end{align}
Each alignment has a conformance cost of $c(\gamma^{\mathrm{opt}})=1$, reflecting a single model move.

While the optimal alignment can be identified directly from the synchronous product in small illustrative examples, this task becomes computationally challenging for larger and more complex models. In practice, the search for an optimal alignment is therefore performed on the reachability graph of the synchronous product, which we define next.

\begin{definition}[Reachability Graph of a Synchronous Product]\label{def:rg_sp}
Let 
\[
SP = (P, T, F, \lambda, m_i, m_f)
\]
be the synchronous product from Definition~\ref{def:synch_prod}, and let
\(c:T \to \mathbb{R}_{\ge 0}\) be the cost function from Definition~\ref{def:cost_func}.
Denote by \(W^-, W^+ \in \mathbb{Z}_{\ge 0}^{|P|\times|T|}\) the \emph{backward} and \emph{forward} incidence matrices,  
where \(W^-(p,t)\) and \(W^+(p,t)\) give, respectively, the number of tokens
consumed from and produced in place \(p\) by firing transition \(t\).
The space \(\mathbb{Z}_{\ge 0}^{|P|\times|T|}\) denotes matrices of nonnegative integers of
dimension \(|P|\times|T|\).

Let the place-transition incidence matrix be
\[
I = W^+ - W^- \qquad (\text{denoted } I_{SP} \text{ when referring to } SP), 
\]
where \(I_{SP}(\cdot,t)\) denotes the column of \(I_{SP}\) corresponding to transition \(t\).

The \emph{reachability graph} of \(SP\) is the directed, edge-labeled graph
\[
\RG(SP) = (V, E, \ell_E, w, m_i, m_f),
\]
defined as follows:
\begin{description}[itemsep=0.1em, topsep=0pt]
  \item \(V \subseteq \mathbb{Z}_{\ge 0}^{|P|}\) is the set of all markings reachable from the initial marking \(m_i\).
  \item \(E \subseteq V \times T \times V\) contains an edge \((m,t,m')\) if and only if \(t\) is enabled at \(m\) and its firing yields \(m'\), that is,
  \[
    \forall p \in P:\; m(p)\ge W^-(p,t),
  \]
  and the successor marking is
  \[
  m' = m + I_{SP}(\cdot,t)
= m - W^-(\cdot,t) + W^+(\cdot,t).
  \]
  \item \(\ell_E: E \to (A \cup \{\tau,\gg\}) \times (A \cup \{\gg\})\) labels each edge by the transition label of \(SP\):
        
        \[
        \ell_E(m,t,m') = \lambda(t),
        \]
        where \(\lambda\) is the labeling function defined in Definition~\ref{def:synch_prod}. 
        
        Hence:
        \begin{description}[itemsep=0.1em, topsep=0pt]
          \item synchronous moves have labels \((a,a)\) for \(a \in A\),
          \item model moves including silent moves have labels \((a,\gg)\) for \(a \in A \cup \{\tau\}\),
          \item log moves have labels \((\gg,a)\) for \(a \in A\).
        \end{description}
  \item \(w:E \to \mathbb{R}_{\ge 0}\) assigns edge weights by the move cost:
        \[
        w(m,t,m') = c(t).
        \]
\end{description}
\vspace{5pt}

An alignment is a finite path 
\[
\pi = m_i \xrightarrow{t_1} m_1 \xrightarrow{t_2} \cdots \xrightarrow{t_k} m_k
\]
in \(\RG(SP)\), where $m_k=m_f$. It induces the transition sequence \newline
\(\gamma = \langle t_1, \dots, t_k \rangle\)
in \(SP\) with total cost
\[
w(\pi) = \sum_{i=1}^{k} w(m_{i-1}, t_i, m_i) = \sum_{i=1}^{k} c(t_i).
\]
An \emph{optimal alignment} corresponds to a minimum-cost alignment path in \(\RG(SP)\).

\end{definition}

\vspace{1ex}

\begin{figure}[ht]
\centering
\includegraphics[width=0.85\textwidth]{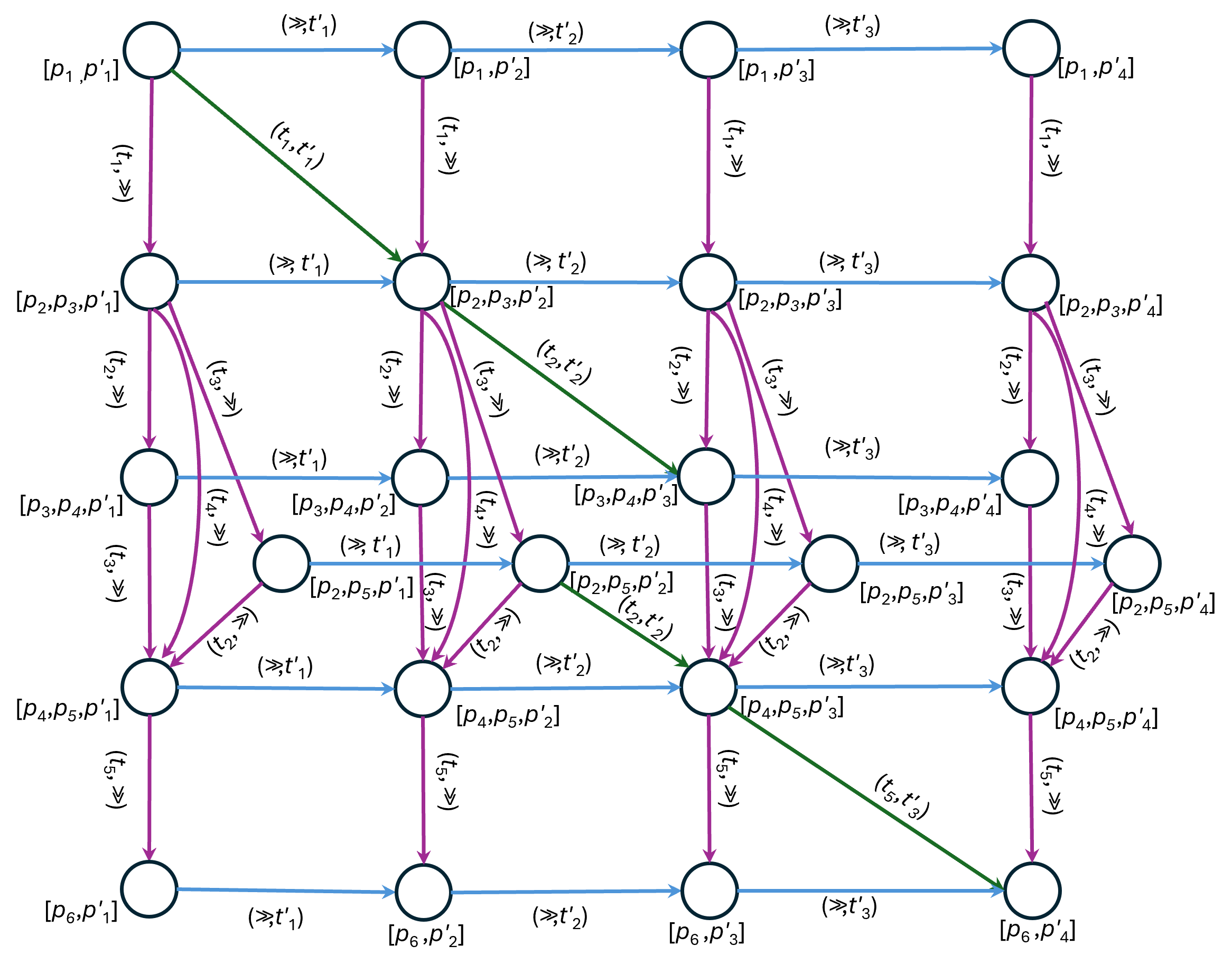}
\caption{The complete reachability graph of the synchronous product in Figure~\ref{fig_synch_prod}. Arcs in blue, green and purple signify log-, synchronous-, and model-moves, respectively.}
\label{fig_rg_sp}
\end{figure}

The reachability graph of the synchronous product in Figure~\ref{fig_synch_prod}, constructed according to Definition~\ref{def:rg_sp}, is shown in Figure~\ref{fig_rg_sp}. Under the standard cost structure, the minimum-cost paths from the initial to the final marking correspond to the two optimal alignments shown in Equation~\eqref{eq:opt_alignment}, each with total cost one.

To motivate our reformulation of the conformance checking problem as a linear program, we begin with a fundamental principle from linear programming theory---the concept of total unimodularity (TU), defined below. Total unimodularity ensures that certain linear programs, when defined over integral right-hand sides, yield integer optimal solutions without explicitly enforcing binary constraints \citep{schrijver1998theory,bertsimas1997introduction}.
This property provides the theoretical foundation for our exploration:
if the constraint matrix underlying the conformance checking formulation is totally unimodular, then the problem can be solved exactly through a linear relaxation.
In \model, we leverage this insight as the key motivation for developing an LP-based reformulation
that preserves the optimal alignment structure while avoiding integer constraints.

\begin{definition}[Total Unimodularity]
A matrix \( A \) is said to be totally unimodular if every square submatrix of \( A \) has a determinant of \( 0 \), \( +1 \), or \( -1 \). That is, for any square submatrix \( B \) of \( A \), the determinant satisfies:
\[
\det(B) \in \{0, \pm1\}.
\]
If the constraint matrix of a linear program (LP) is totally unimodular and the right-hand side vector is integral, then every extreme point of the feasible region is integral, ensuring that the LP relaxation of an integer linear program has an integer optimal solution.
\end{definition}

\begin{definition}[LP Solution being Optimal for the MILP]
Consider an integer linear program:
\[
\min \; c^T x \quad \text{subject to } Ax \leq b, \; x \geq 0, \; x \in \mathbb{Z}^n.
\]
Its linear programming (LP) relaxation is obtained by allowing \(x\) to take real values:
\[
\min \; c^T x \quad \text{subject to } Ax \leq b, \; x \geq 0, \; x \in \mathbb{R}^n.
\]

If the feasible region of the LP is an integral polyhedron (i.e., all its vertices are integral), then every optimal solution of the LP relaxation is also an optimal solution of the MILP. A sufficient condition for this integrality is that the constraint matrix \(A\) is totally unimodular and the right-hand side vector \(b\) is integral.
\end{definition}

\section{Model Development}\label{sec:model}

Multiple previous studies have shown that searching for an optimal alignment using $A^*-$like algorithms with various heuristics, or solving a Mixed-Integer Linear Programming (MILP) formulation of the problem, can be time consuming. We propose an exact reformulation of the alignment problem as a minimum-cost flow LP on a network. Because the node-arc incidence matrix of this network is totally unimodular and the right‐hand side is integral, the linear programming relaxation is integral. Therefore, for a given constructed reachability graph, the resulting optimization problem is polynomially solvable in the size of that network \citep{tardos1986strongly,schrijver2003combinatorial}.

Having defined the synchronous product \(SP=(P,T,F,\lambda,m_i,m_f)\),
its reachability graph \(\RG(SP)\), and the cost structure \(c:T\to\mathbb{R}_{\ge0}\),
we now formulate the alignment-based conformance checking problem as a
minimum-cost network flow problem.
This reformulation preserves the structure of optimal alignments while enabling
a totally unimodular (TU) linear programming representation that is solvable in polynomial time.

\subsection{From Reachability Graph to Network Flow}

Let \(\RG(SP)=(V,E,\ell_E,w,m_i,m_f)\) denote the reachability graph of the synchronous product
(Definition~\ref{def:rg_sp}).  
Each node \(v\in V\) corresponds to a reachable marking of \(SP\),
and each directed edge \(e=(v,t,v')\in E\) represents the firing of transition \(t\in T\)
from marking \(v\) to marking \(v' = v + I_{SP}(\cdot,t)\).
The weight function \(w:E\to\mathbb{R}_{\ge0}\) assigns the move cost \(w(e)=c(t)\).

The problem of finding an optimal alignment now becomes that of determining
a minimum-cost path in \(\RG(SP)\)
from the initial to the final marking.

\subsection{Decision Variables and Flow Conservation} \label{sec:decision variables}

For every edge \(e\in E\), introduce a continuous decision variable
\[
x_e \in [0,1],
\]
representing the proportion of \emph{case flow} traversing edge \(e\).
In integral optimal solutions, \(x_e=1\) indicates that the corresponding
transition is used in the optimal alignment, and \(x_e=0\) otherwise.

Define the \emph{balance vector} \(b\in\mathbb{Z}^{|V|}\) by
\[
b_v =
\begin{cases}
  1, & v=m_i,\\[2pt]
 -1, & v=m_f,\\[2pt]
  0, & \text{otherwise}.
\end{cases}
\]
Flow conservation is then expressed by
\begin{equation}\label{eq:flowcons}
\sum_{e\in\delta^+(v)} x_e - \sum_{e\in\delta^-(v)} x_e = b_v,
\qquad \forall v\in V,
\end{equation}
where \(\delta^+(v)\) and \(\delta^-(v)\) denote the sets of outgoing and incoming edges at node \(v\),
respectively.  
This enforces one unit (i.e., a single case) of flow from the initial to the final marking, with zero net flow elsewhere.

\subsection{Linear Programming Formulation}

The alignment problem can now be written as the following linear program:
\begin{align}
\min_{x} \quad & \sum_{e\in E} w(e)\,x_e \label{eq:lpobj}\\
\text{s.t.}\quad
& \sum_{e\in\delta^+(v)} x_e - \sum_{e\in\delta^-(v)} x_e = b_v,
  && \forall v\in V, \label{eq:lpflow}\\
& 0 \le x_e \le 1, && \forall e\in E. \label{eq:lpbounds}
\end{align}

The objective~\eqref{eq:lpobj} accumulates the total move cost along a chosen path.
Constraints~\eqref{eq:lpflow} and~\eqref{eq:lpbounds} ensure conservation of flow
and limit the feasible region to unit flow between \(m_i\) and \(m_f\).
An optimal basic feasible solution to this LP corresponds to a path in \(\RG(SP)\)
whose total cost equals the minimal conformance deviation.

\subsection{Total Unimodularity and Integrality}

Let \(B\in\{-1,0,1\}^{|V|\times|E|}\) be the node--arc incidence matrix of the \textit{reachability graph} \(\RG(SP)\),
where \(B_{v,e}=1\) if \(e\in\delta^+(v)\), \(B_{v,e}=-1\) if \(e\in\delta^-(v)\), and \(0\) otherwise.
Constraints~\eqref{eq:lpflow} can thus be written compactly as
\[
Bx = b.
\]
Since \(B\) is the incidence matrix of a directed network,
it is totally unimodular \citep{tardos1986strongly,schrijver2003combinatorial}. For completeness, the full incidence matrix of the reachability graph in Figure~\ref{fig_rg_sp}, is provided in the supplementary material.

This follows directly from the fact that each column of $B$ contains exactly one $+1$ and one $-1$, with all other entries equal to zero.
Because \(b\) is integral and the variable bounds in Constraints~\eqref{eq:lpbounds} are integral,
every basic feasible solution of the LP is integral (\(x_e\in\{0,1\}\)).
Therefore, solving the LP yields an optimal integral path
without requiring explicit binary constraints.

\begin{remark}[Interpretation]
The TU property implies that the LP formulation~\eqref{eq:lpobj}--\eqref{eq:lpbounds}
is equivalent to its integer version.
A resulting solution corresponds to a single \(m_i\!\to m_f\) path in \(\RG(SP)\),
representing an optimal alignment \(\gamma^{\mathrm{opt}}\) of the synchronous product.
Hence, \model\ achieves exact alignment computation through a purely linear formulation.
\end{remark}

\subsection{Recovering the Optimal Alignment}

Let \(\mathcal{P} = \{ e_1, e_2, \dots, e_k \}\) denote the set of edges with \(x_{e_i}=1\)
in the optimal solution.
Traversing \(\mathcal{P}\) reconstructs the optimal firing sequence
\(\gamma^{\mathrm{opt}} = \langle t_1, t_2, \dots, t_k \rangle\),
where each \(e_i=(v_{i-1},t_i,v_i)\).

Each transition in the synchronous product has two labels, one from the process model side and one from the trace side. Hence, the labeling function returns a pair:
\[
\lambda: T \to (A \cup \{\tau, \gg\}) \times (A \cup \{\gg\}),
\]
where $(a,a)$ is a synchronous move, $(a,\gg)$ for $a \in A \cup \{\tau\}$ is a model move, and $(\gg,a)$ for $a \in A$ is a log move. The total alignment cost is \(c(\gamma^{\mathrm{opt}}) = \sum_i c(t_i)\), which equals the LP objective value.

\subsection{Complexity and Scalability}

Since the constraint matrix \(B\) is TU, the 
problem~\eqref{eq:lpobj}--\eqref{eq:lpbounds} is integral.
Therefore, for a given explicitly constructed reachability graph,
the resulting optimization problem is polynomially solvable in the size of
that graph and yields exact alignments without introducing integer variables.
This remains valid for synchronous products with loops.

In practical terms, \model\ replaces the traditional $A^*$ or MILP search
with a single LP of size \((|E|,|V|)\),
whose solution directly yields the optimal alignment and conformance cost.

Note that this does not contradict the PSPACE-hardness of reachability, since the reachability graph itself may be exponentially large.

\subsection{MILP Formulation on the Synchronous Product}\label{sec:milp-sp}

An alternative, straightforward, mathematical programming formulation of the alignment-based conformance checking problem can be derived directly on the synchronous product, rather than on its (typically much larger) reachability graph. However, as will be shown, the main drawback of this formulation is that its constraint matrix does not, in general, possess the total unimodularity property, thereby requiring the problem to be solved as a computationally hard integer linear program.

Let \(SP=(P,T,F,\lambda,m_i,m_f)\) be the synchronous product (Definition~\ref{def:synch_prod}).
Denote by \(I \equiv I_{SP}\in\mathbb{Z}^{|P|\times |T|}\) the place–transition incidence matrix of \(SP\),
by \(c\in\mathbb{R}_{\ge 0}^{|T|}\) the move-cost vector (cf. Definition~\ref{def:cost_func}),
and by \(m_i,m_f\in\mathbb{Z}_{\ge 0}^{|P|}\) the initial and final markings.
Fix an upper bound \(n\in\mathbb{Z}_{>0}\) on the alignment length.

The decision variables \(x_{j,k}\in\{0,1\}\) that equals 1 if transition \(j\in T\) fires at step \(k\in\{1,\dots,n\}\), and \(z_k\in\{0,1\}\) that equals 1 if the alignment has terminated at or before step \(k\). For convenience, let $x_{\cdot,k} := (x_{1,k},\dots,x_{|T|,k})^\top$ denote the
vector of transition-decision variables at step $k$.

Mathematically, we formulate the alignment-based optimization problem as:
\begin{align}
\min_{x,z} \quad & \sum_{k=1}^{n}\sum_{j=1}^{|T|} c_j\,x_{j,k}
\label{eq:milp-obj}\\
\text{s.t.}\quad 
& m_i + I\!\Big(\sum_{k=1}^{n} x_{\cdot,k}\Big) = m_f,
\label{eq:milp-fm}\\[3pt]
& m_i + I\!\Big(\sum_{\tau=1}^{k} x_{\cdot,\tau}\Big) \ge 0,
\quad \forall k=1,\dots,n,
\label{eq:milp-prefix}\\[3pt]
& \sum_{j=1}^{|T|} x_{j,k} + z_k = 1,
\quad \forall k=1,\dots,n,
\label{eq:milp-one}\\[3pt]
& z_{k+1} \ge z_k,
\quad \forall k=1,\dots,n-1,
\label{eq:milp-mono}\\[3pt]
& x_{j,k}\in\{0,1\},\; z_k\in\{0,1\}.
\label{eq:milp-dom}
\end{align}

The objective, \eqref{eq:milp-obj}, minimizes the total cost of all fired transitions. Synchronous moves incur zero cost, visible log/model deviations cost 1, and \(\tau\)-model moves incur a negligible cost \(\epsilon>0\) (e.g., $10^{-6}$).

Constraint \eqref{eq:milp-fm} ensures global token balance across all places, requiring that the fired transitions transform the initial marking \(m_i\) into the final marking \(m_f\).
Constraint \eqref{eq:milp-prefix} guarantees that all intermediate markings remain nonnegative, thereby preserving Petri-net firing feasibility at every prefix of the alignment.
Constraint \eqref{eq:milp-one} enforces that at each step exactly one transition fires or, alternatively, that the alignment has already terminated.

Constraint \eqref{eq:milp-mono} ensures that once the alignment terminates (\(z_k=1\)), it remains terminated in all subsequent steps.

Binary decision variables are defined by Constraint \eqref{eq:milp-dom} for all transition-step combinations and termination indicators. 

A natural LP relaxation replaces \eqref{eq:milp-dom} by
\(
0\le x_{j,k}\le 1,\; 0\le z_k\le 1
\),
while keeping \eqref{eq:milp-fm}--\eqref{eq:milp-mono} unchanged. This provides a lower bound for the MILP and can serve as a heuristic but \textit{it does not guarantee integrality}.
This is because the incidence matrix \(I\) is not totally unimodular for many Petri-nets that include free-choice, AND-splits and joins, synchronization and loops. Even if the incidence matrix would be totally unimodular, the combined constraint matrix in
\eqref{eq:milp-fm}--\eqref{eq:milp-mono} is typically \emph{not} totally unimodular since the prefix-feasibility constraints \eqref{eq:milp-prefix} introduce cumulative blocks that place multiple identical copies of \(I\) in the same columns across many rows. Moreover, the per-step equalities \eqref{eq:milp-one} couple all \(x_{\cdot,k}\) densely within each time slice; and 
these dense assignment-like rows interact with the global balance rows \eqref{eq:milp-fm}.
Together, these patterns violate standard TU criteria (e.g., Ghouila-Houri), so the LP relaxation can admit fractional solutions in general, which was confirmed in experiments.

Thus, the LP relaxation of the MILP is generally insufficient, necessitating the solution of the full MILP to obtain an optimal alignment. As expected, preliminary experiments confirmed that MILP formulations consistently exceeded practical time limits and are therefore excluded from our empirical evaluation.

\section{Illustrative Example as a Minimum-Cost Network Flow}\label{sec:example-mcf}

\begin{figure*}[h!t]
\centering
\begin{subfigure}[b]{0.56\textwidth}
    \centering
    \includegraphics[width=\textwidth]{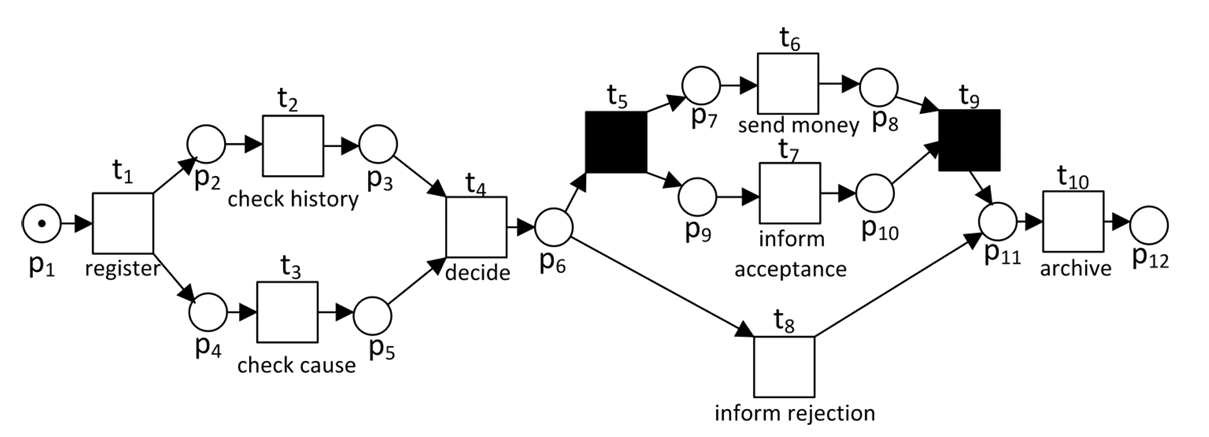}
    \caption{}
    \label{fig4_1_Adriansyah}
\end{subfigure}
\hfill
\begin{subfigure}[b]{0.43\textwidth}
    \centering
    \includegraphics[width=\textwidth]{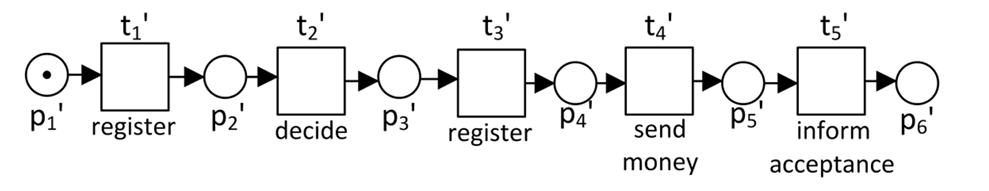}
    \caption{}
    \label{fig4_2_Adriansyah}
\end{subfigure}\\
[0.5em]
\begin{subfigure}[b]{0.85\textwidth}
    \centering
    \includegraphics[width=\textwidth]{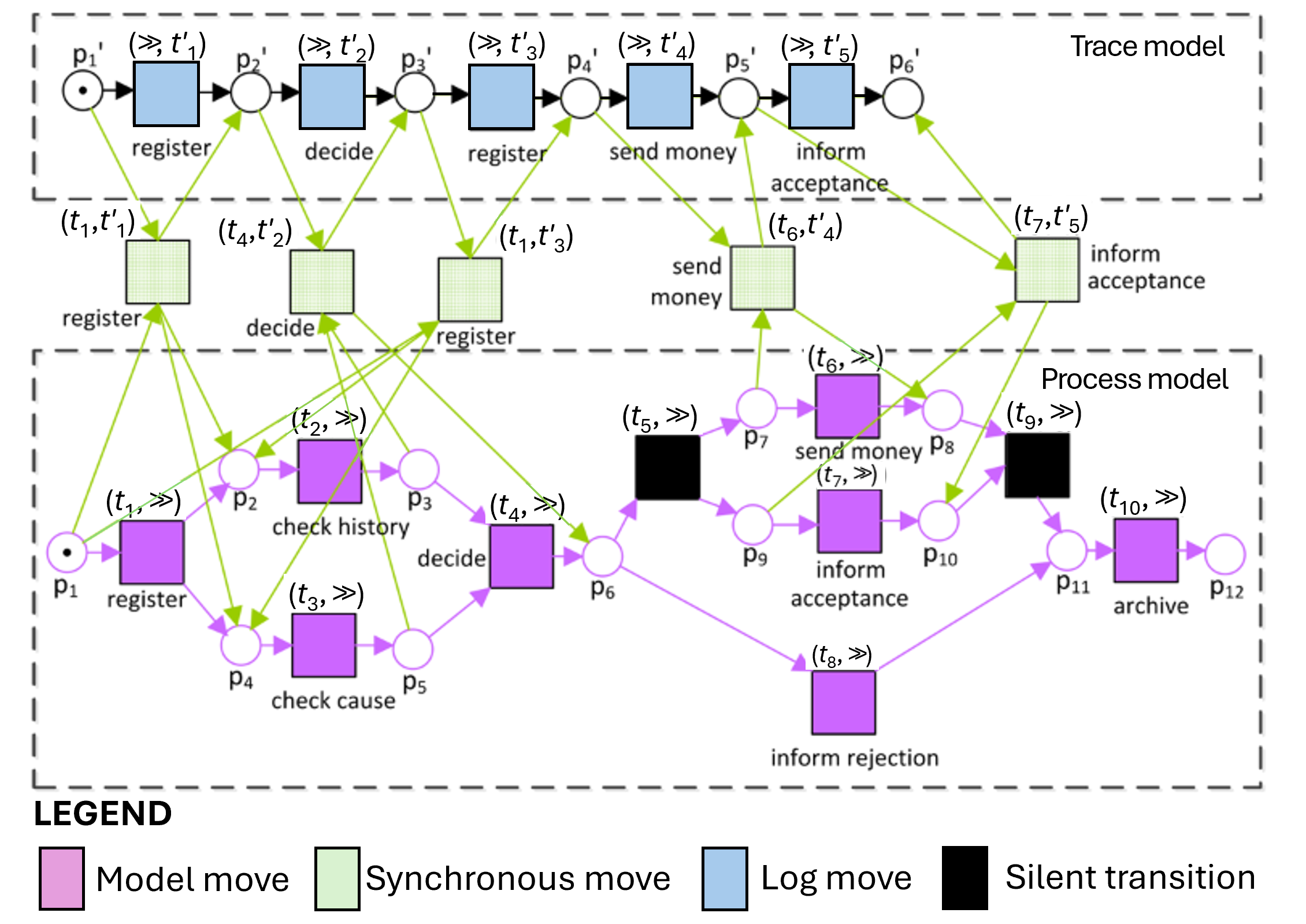}
    \caption{}
    \label{fig4_3_Adriansyah}
\end{subfigure}
\caption{(a) A process of handling insurance claims by an insurance company, and (b) a trace representing a process realization. (c) is the Synchronous product between the trace (labeled as an event net) and the process model (labeled as a process net) with synchronous moves. The three figures were taken from \citet{Adriansyah2014}.}
\label{fig_process_trace_Adriansh}
\end{figure*}

We now use the illustrative example shown in Fig.~\ref{fig_process_trace_Adriansh} (process, trace, and their synchronous product) to demonstrate how the LP formulation can be instantiated as a minimum-cost flow problem on the state graph induced by the synchronous product.

\subsection{State Graph from the Synchronous Product}
Let $SP=(P,T,F,\lambda,m_i,m_f)$ be the synchronous product in Fig.~\ref{fig_process_trace_Adriansh}c.  
Its \emph{state graph} is $G=(V,E)$ with
\begin{description}
  \item $V=\mathcal{R}(SP)$: all markings reachable from $m_i$ in $SP$, and
  \item $E=\{\,e=(v,t,v') \mid v\in V,\ t\in T \text{ enabled at } v,\ v'=v+I_{SP}(\cdot,t)\,\}$, where $I_{SP}$ is the incidence matrix of the synchronous product (see, Table~\ref{tab:unimodular}).
\end{description}
For each $v\in V$ we write the outgoing and incoming arc sets as
\[
\begin{aligned}
\delta^+(v) &= \{\, e=(v,t,v') \in E \,\},\\
\delta^-(v) &= \{\, e=(u,t,v) \in E \,\}.
\end{aligned}
\]
Enabledness is defined using the backward incidence matrix $W^-$ of the synchronous product $SP$---a transition $t$ is enabled at marking $v$ if and only if $v(p)\ge W^-(p,t)$ for all $p\in P$.

\begin{table*}[ht]
\centering
\small
\setlength{\tabcolsep}{3pt}
\caption{Incidence matrix of the synchronous product of Fig.~\ref{fig_process_trace_Adriansh}.}
\label{tab:unimodular}
\resizebox{0.95\textwidth}{!}{%
\begin{tabular}{@{}l|*{5}{c}|*{10}{c}|*{5}{c}@{}}
\toprule
& \multicolumn{5}{c|}{\textbf{Synchronous Moves}} & \multicolumn{10}{c|}{\textbf{Model Moves}} & \multicolumn{5}{c}{\textbf{Log Moves}} \\
& \rotatebox{90}{$(t_1,t_1')$} 
& \rotatebox{90}{$(t_4,t_2')$} 
& \rotatebox{90}{$(t_1,t_3')$} 
& \rotatebox{90}{$(t_6,t_4')$} 
& \rotatebox{90}{$(t_7,t_5')$} 
& \rotatebox{90}{$(t_1,\gg)$} 
& \rotatebox{90}{$(t_2,\gg)$} 
& \rotatebox{90}{$(t_3,\gg)$} 
& \rotatebox{90}{$(t_4,\gg)$} 
& \rotatebox{90}{$(t_5,\gg)$} 
& \rotatebox{90}{$(t_6,\gg)$} 
& \rotatebox{90}{$(t_7,\gg)$} 
& \rotatebox{90}{$(t_8,\gg)$} 
& \rotatebox{90}{$(t_9,\gg)$} 
& \rotatebox{90}{$(t_{10},\gg)$} 
& \rotatebox{90}{$(\gg,t_1')$} 
& \rotatebox{90}{$(\gg,t_2')$} 
& \rotatebox{90}{$(\gg,t_3')$} 
& \rotatebox{90}{$(\gg,t_4')$} 
& \rotatebox{90}{$(\gg,t_5')$} \\
\midrule
$p_1$ & $-1$ & 0 & $-1$ & 0 & 0 & $-1$ & 0 & 0 & 0 & 0 & 0 & 0 & 0 & 0 & 0 & 0 & 0 & 0 & 0 & 0 \\
$p_2$ & 1 & 0 & 1 & 0 & 0 & 1 & $-1$ & 0 & 0 & 0 & 0 & 0 & 0 & 0 & 0 & 0 & 0 & 0 & 0 & 0 \\
$p_3$ & 0 & $-1$ & 0 & 0 & 0 & 0 & 1 & 0 & $-1$ & 0 & 0 & 0 & 0 & 0 & 0 & 0 & 0 & 0 & 0 & 0 \\
$p_4$ & 1 & 0 & 1 & 0 & 0 & 1 & 0 & $-1$ & 0 & 0 & 0 & 0 & 0 & 0 & 0 & 0 & 0 & 0 & 0 & 0 \\
$p_5$ & 0 & $-1$ & 0 & 0 & 0 & 0 & 0 & 1 & $-1$ & 0 & 0 & 0 & 0 & 0 & 0 & 0 & 0 & 0 & 0 & 0 \\
$p_6$ & 0 & 1 & 0 & 0 & 0 & 0 & 0 & 0 & 1 & $-1$ & 0 & 0 & $-1$ & 0 & 0 & 0 & 0 & 0 & 0 & 0 \\
$p_7$ & 0 & 0 & 0 & $-1$ & 0 & 0 & 0 & 0 & 0 & 1 & $-1$ & 0 & 0 & 0 & 0 & 0 & 0 & 0 & 0 & 0 \\
$p_8$ & 0 & 0 & 0 & 1 & 0 & 0 & 0 & 0 & 0 & 0 & 1 & 0 & 0 & $-1$ & 0 & 0 & 0 & 0 & 0 & 0 \\
$p_9$ & 0 & 0 & 0 & 0 & $-1$ & 0 & 0 & 0 & 0 & 1 & 0 & $-1$ & 0 & 0 & 0 & 0 & 0 & 0 & 0 & 0 \\
$p_{10}$ & 0 & 0 & 0 & 0 & 1 & 0 & 0 & 0 & 0 & 0 & 0 & 1 & 0 & $-1$ & 0 & 0 & 0 & 0 & 0 & 0 \\
$p_{11}$ & 0 & 0 & 0 & 0 & 0 & 0 & 0 & 0 & 0 & 0 & 0 & 0 & 1 & 1 & $-1$ & 0 & 0 & 0 & 0 & 0 \\
$p_{12}$ & 0 & 0 & 0 & 0 & 0 & 0 & 0 & 0 & 0 & 0 & 0 & 0 & 0 & 0 & 1 & 0 & 0 & 0 & 0 & 0 \\
\midrule
$p_1'$ & $-1$ & 0 & 0 & 0 & 0 & 0 & 0 & 0 & 0 & 0 & 0 & 0 & 0 & 0 & 0 & $-1$ & 0 & 0 & 0 & 0 \\
$p_2'$ & 1 & $-1$ & 0 & 0 & 0 & 0 & 0 & 0 & 0 & 0 & 0 & 0 & 0 & 0 & 0 & 1 & $-1$ & 0 & 0 & 0 \\
$p_3'$ & 0 & 1 & $-1$ & 0 & 0 & 0 & 0 & 0 & 0 & 0 & 0 & 0 & 0 & 0 & 0 & 0 & 1 & $-1$ & 0 & 0 \\
$p_4'$ & 0 & 0 & 1 & $-1$ & 0 & 0 & 0 & 0 & 0 & 0 & 0 & 0 & 0 & 0 & 0 & 0 & 0 & 1 & $-1$ & 0 \\
$p_5'$ & 0 & 0 & 0 & 1 & $-1$ & 0 & 0 & 0 & 0 & 0 & 0 & 0 & 0 & 0 & 0 & 0 & 0 & 0 & 1 & $-1$ \\
$p_6'$ & 0 & 0 & 0 & 0 & 1 & 0 & 0 & 0 & 0 & 0 & 0 & 0 & 0 & 0 & 0 & 0 & 0 & 0 & 0 & 1 \\
\bottomrule
\end{tabular}
}
\end{table*}

\subsection{Move Costs for the Example}
We adopt the move-cost function of Definition~\ref{def:cost_func} by which synchronous moves have cost $0$, $\tau$-model moves have cost $\epsilon > 0$ (tiny), and all other non-synchronous moves have cost $1$.  
For Fig.~\ref{fig_process_trace_Adriansh}, the $|T|=20$ transitions are ordered as in Table~\ref{tab:unimodular}, with the cost vector
\[
c=(0,0,0,0,0,1,1,1,1,\epsilon,1,1,1,\epsilon,1,1,1,1,1,1),
\]
where index $j$ corresponds to the $j$-th transition (column) of Table~\ref{tab:unimodular}.  
Each arc $e=(v,t_j,v')\in E$ inherits the transition cost, i.e., $w(e)=c_j$.

\subsection{Start/End Balance}
For Fig.~\ref{fig_process_trace_Adriansh}, the initial marking is $m_i = [p_1,p'_1]$ and the final marking is $m_f = [p_{12},p_6']$. The balance vector $b$ routes a single case from $m_i$ to $m_f$ as defined in Section~\ref{sec:decision variables}.

\subsection{Min-Cost Flow LP}
Let $x_e\in[0,1]$ be the flow on arc $e\in E$. The alignment problem is an instance of the LP formulation~\eqref{eq:lpobj}--\eqref{eq:lpbounds}, with edge weights $w(e)$ as defined in Definition~\ref{def:rg_sp} and balance vector $b$ enforcing a single unit of flow from $m_i$ to $m_f$.

\subsection{Solving and Reading the Alignment}
The node-arc incidence matrix of $G$ is totally unimodular; since $b$ is integral, an optimal \emph{basic} solution of the LP is integral and thus yields a single $m_i\!\to m_f$ path $\mathcal{P}=\{e_1,\dots,e_L\}$. 

Traversing $\mathcal{P}$ recovers the alignment. Mathematically, for $e_\ell=(v_\ell,t_\ell,v_{\ell+1})$, append
\begin{description}
  \item $(a,a)$ if $t_\ell$ is a synchronous move on label $a$,
  \item $(a,\gg)$ if $t_\ell$ is a log move on $a$,
  \item $(\gg,a)$ if $t_\ell$ is a model move on $a$ (including $\tau$).
\end{description}
Because synchronous transitions have zero-cost, the optimal path uses synchronous arcs wherever enabled, inserting the minimum number of (costly) log/model moves necessary to connect $m_i$ to $m_f$.

\section{Experiments}
\label{sec:experiments}

\subsection{Experimental Design}
\label{sec:experimental-design}

Our experimental evaluation, including more than 2.1 million process instances across a variety of publicly available process mining datasets, aims to identify the conditions under which one approach may be more suitable, and the key factors governing algorithms' performance.

Initial experiments revealed that performance varies across datasets in ways not explained by conventional factors such as trace length or model size. We hypothesize that fitness affects the depth of the search methods such as $A^*$ (that is, the number of alignment steps), and precision governs the size of the synchronous product and reachability graph explored in optimization-based approaches.

Therefore, for each dataset we discovered a collection of process models exhibiting varying levels of fitness and precision by applying several well-established discovery algorithms, including the Alpha Miner~\citep{van2004workflow}, the Heuristics Miner using multiple dependency-filtering thresholds~\citep{weijters2006process}, and the Inductive Miner (IM)~\citep{leemans2013discovering}, for which we performed a noise-threshold search over the range $[0.0, 0.4]$. Fitness and precision were evaluated using token-based replay techniques following \cite{rozinat2008conformance}.
We selected process models using a two-tier strategy. For most datasets, we targeted sound models achieving fitness $\ge 0.9$ and precision $\ge 0.7$, consistent with quality thresholds commonly employed in real-world process mining applications. To test our hypothesis that alignment cost proxied by model fitness is the primary determinant of LP advantage, we supplemented these models with additional variants spanning the fitness-precision spectrum. This design enables systematic comparison of algorithm performance across diverse model characteristics. For the PLG2 synthetic benchmark, we employed the reference models provided with each generated log~\citep{burattin2010plg,munoz2014single}. We also screened additional candidate logs, including BPI Challenge 2015, but did not include them when the discovered process models consistently exhibited very low precision across discovery methods, since such underconstrained models are not suitable reference models for the type of control-flow conformance evaluation considered in this paper.

By comparing algorithm behavior across a fitness-precision spectrum within each dataset, we can isolate systematic empirical relationships between model characteristics and the relative efficiency of $A^*$ versus LP, independent of dataset-specific confounds.

In the used LP implementation, we first construct explicitly a bounded reachability subgraph of the synchronous product by breadth-first exploration up to a practical depth limit, and only then solve the LP on that constructed subgraph. Hence, unless stated otherwise, the reported RG node counts, edge counts, and RG-construction times refer to this bounded subgraph. Instances
for which graph construction exceeded the time or size limit were reported as timeouts. For all instances solved by both methods, the LP objective value matched the $A^*$ objective value, as expected.
In the implementation, $\tau$-edge pruning was disabled, and the only active pruning during reachability-graph construction was elimination of self-loops, which is safe because self-loops leave the marking unchanged and, under the nonnegative move-cost structure used here, cannot improve a minimum-cost path to the final marking.

\subsection{Hardware and Software Environment}
Experiments were executed on a server with two Intel Xeon Gold 6248R processors (48 cores) and 403\,GB RAM. All code ran in \texttt{Python} 3.9.21. Both methods were executed sequentially on a single process.
 
The two methods rely on LP solvers for different purposes and at different scales. The URC$^2$ formulation was solved using \texttt{Gurobi}~11 (primal simplex method, 30-second timeout). \texttt{PM4Py}'s $A^*$ implementation (version 2.7.19.1) uses \texttt{cvxopt\_solver\_custom\_align}, a lightweight LP backend optimized for alignment computation, to solve the marking-equation heuristic during search. Each heuristic LP has $O(|P|+|T|)$ variables and is invoked at every expanded search state. In contrast, URC$^2$ solves a single LP defined over the reachability graph, with a much higher number of variables that corresponds to the number of edges in the constructed RG. We verified that this solver asymmetry does not favor URC$^2$. On the contrary, replacing \texttt{PM4Py}'s solver with Gurobi for the marking-equation heuristic resulted in a significant (\>3$\times$) slowdown of $A^*$, because Gurobi's per-call overhead dominates on the small LPs called hundreds of times per trace. The default solver configuration is therefore already favorable to $A^*$.

We compare against standard $A^*$ rather than specialized variants because
(1) it is the default in major toolkits and represents the highly optimized,
practitioner-facing implementation most commonly used in practice;
(2) advanced variants improve heuristics and pruning within the same search
paradigm but do not change the fundamental sensitivity of best-first search
to alignment cost, and (3) our contribution, polynomial-time optimality via LP, is independent of the baseline.

\texttt{PM4Py} benefits from CPython-compiled internals (e.g., the heap queue
implemented in~C and a custom-optimized \texttt{cvxopt} backend) and years of
engineering optimization. URC$^2$'s reachability graph construction accelerates
the core successor-computation loops via Numba JIT compilation, but the
surrounding orchestration remains in Python. Overall, \texttt{PM4Py} represents a substantially more mature and optimized implementation, which if anything favors~$A^*$ in the reported comparison.
To support reproducibility, the implementation and experiment scripts are publicly available in the author's
\href{https://github.com/Izack-Cohen/unimodular-conformance-checking}{GitHub repository}. Raw event logs are not redistributed and should be obtained from their original public sources.

\subsection{Datasets}

To evaluate the conformance checking approaches, we selected four widely studied real-life
event logs from different domains--public administration, financial services, and healthcare, and a set of standard synthetic datasets from the PLG2 process generator with a reference model. These datasets are typical in process mining research and provide a broad spectrum of trace lengths, behavioral complexity, and model sizes. Table~\ref{tab:datasets_overview} summarizes key
characteristics of the used datasets. 

\begin{table}[htbp]
\centering
\small
\setlength{\tabcolsep}{4pt}
\caption{Overview of the datasets used in our study.}
\label{tab:datasets_overview}
\resizebox{0.95\textwidth}{!}{%
\begin{tabular}{@{}p{3.8cm}lcp{4.2cm}@{}}
\toprule
\textbf{Dataset} & \textbf{Domain} & \textbf{Length} & \textbf{Notes} \\
\midrule
Road Traffic Fine \citep{deleoni2015road} 
  & Public Administration
  & 2--20 
  & High-volume, low variability \\[0.5em]
BPI Challenge 2012 \citep{vandongen2012bpi2012} 
  & Financial Services
  & 3--175 
  & Loan application with complex decision points \\[0.5em]
BPI Challenge 2013 \citep{steeman2013bpi2013} 
  & IT Service Management
  & 1--123 
  & Help-desk tickets; low structural complexity \\[0.5em]
Sepsis Cases \citep{mannhardt2016sepsis} 
  & Healthcare 
  & 1--400 
  & High variability; emergency workflows \\[0.5em]
PLG2 Synthetic \citep{burattin2010plg} 
  & Benchmark 
  & 12--167 
  & Synthetic models with noise injection \\
\bottomrule
\end{tabular}
}
\end{table}

Here is a brief description of each dataset.

\paragraph{Road Traffic Fine Management Process}
This event log originates from an Italian municipal system for managing traffic fines. It contains real-world cases with short traces (2--20 events) representing
procedural steps such as fine creation, payment, and appeal. The dataset is frequently used in
benchmarking studies due to its high volume and simple behavioral structure~\cite{deleoni2015road}.

\paragraph{BPI Challenge 2012 (Loan Application)}
The BPI Challenge 2012 event log~\citep{vandongen2012bpi2012} corresponds to a Dutch financial
institution's loan application process. It contains traces with more
complex structures than the Road traffic Fine Management event log, including data attributes, cancellations, and loops. With traces ranging from 3 to 175 activities. It is widely used in alignment, discovery, and performance analysis research.

\paragraph{BPI Challenge 2013 (Incident Management)}
This dataset captures an incident and problem management process
from a large IT service provider~\citep{steeman2013bpi2013}. Trace lengths range range from 1--123 events, and the process has a low structural complexity with discovered models that typically contain less than 20 places. The dataset has been used extensively for evaluating conformance checking, predictive monitoring, and behavior analysis.

\paragraph{Sepsis Cases (Hospital Emergency Care)}
The Sepsis Cases log~\citep{mannhardt2016sepsis} contains patient event recordings from the
emergency department of a Dutch hospital. It includes clinical activities such as lab tests,
diagnosis, and treatment steps. The dataset is characterized by high trace variability and
complex medical processes with 3--185 events per case. It is a standard benchmark for
evaluating conformance checking under high variability.

\paragraph{Synthetic PLG2 Benchmark Suite}
We employed synthetic logs generated using the PLG2 process generator~\citep{burattin2010plg}. The benchmark includes models, clean traces (perfect fitness) and noise-augmented variants with injected deviations, enabling controlled comparison of algorithm behavior under varying conformance levels.

\subsection{Performance Metrics}
We evaluated the methods using the following metrics:
\begin{itemize}[label=--,nosep]
    \item \textit{Optimality}: Percentage of traces solved to optimality within the time limit (30 s).
    \item \textit{Cost Agreement}: Verification that both methods produced identical alignment costs was confirmed for all completed instances.
    \item \textit{Computation Time}: Mean across all solved traces.
    \item \textit{Scalability}: Performance correlation with trace length, model size, etc.
\end{itemize}
When relevant, we categorized results by trace length, fitness, and model size to identify where one method begins to outperform the other.

\section{Results}
\label{sec:results}

This section reports the results for the investigated datasets and process–model variants.
Throughout this section, we distinguish between \emph{traces} and \emph{conformance checking instances}.
A conformance checking instance is defined as a pair $(\sigma, M)$, where a trace $\sigma$ is evaluated against a specific process–model variant $M$.
Because multiple model variants are evaluated per dataset, the number of instances substantially exceeds the number of traces.

Unless stated otherwise, all performance statistics (runtime, win rates, speedups) are averages computed over completed conformance checking instances.
The total number of evaluated conformance checking instances exceeds 2.1 million.

\subsection{Aggregate Performance Overview}

Table~\ref{tab:aggregate-results} summarizes the performance of $A^*$ and LP across the datasets. The results confirm that both methods produce identical optimal alignment costs, validating the theoretical correctness of the LP formulation.

\begin{table*}[htbp]
\centering
\setlength{\tabcolsep}{2pt}
\caption{Aggregate performance comparison across datasets. All alignment costs matched between $A^*$ and LP (100\% cost agreement). Win rate indicates the percentage of instances where LP was faster than $A^*$. When Optimal is less than 100\% some instances timed out. 
$\dagger$ PLG2 comprises 4 unique process models evaluated across 21 log configurations (clean/noisy variants with different trace lengths).}
\label{tab:aggregate-results}
\resizebox{\textwidth}{!}{%
\begin{tabular}{@{}lrrr*{3}{r}rrrrc@{}}
\toprule
& & & & \multicolumn{3}{c}{\textbf{Trace Length}} & & \multicolumn{2}{c}{\textbf{Mean Time}} & & \\
\cmidrule(lr){5-7}
\cmidrule(lr){9-10}
\textbf{Dataset} & \textbf{Traces} & \textbf{Instances} & \textbf{Models} & \textbf{Min} & \textbf{Mean} & \textbf{Max} & \textbf{Places} & \textbf{$A^*$} & \textbf{LP} & \textbf{Win} & \textbf{Optimal} \\
& (\#) & (\#) & (\#) & & & & (mean) & (ms) & (ms) & (\%) & (\%) \\
\midrule
Sepsis Cases & 1{,}050 & 9{,}450 & 9 & 3 & 13.9 & 185 & 25.5 & 788 & 1{,}172 & 19.1 & 95.3 \\
BPI Challenge 2012 & 13{,}087 & 52{,}348 & 4 & 3 & 20.0 & 175 & 32.5 & 356 & 128 & 26.1 & 100.0 \\
BPI Challenge 2013 & 7{,}554 & 75{,}540 & 10 & 1 & 8.7 & 123 & 9.7 & 8.6 & 18.1 & 29.5 & 100.0 \\
Road Traffic Fine & 150{,}370 & 1{,}954{,}810 & 13 & 2 & 3.7 & 20 & 20.2 & 9.3 & 113.5 & 2.7 & 100.0 \\
PLG2 Synthetic & 40{,}500 & 40{,}500 & 4$^\dagger$ & 12 & 33.9 & 167 & 37--317 & 630 & 688 & 14.1 & 99.3 \\
\midrule
\textbf{Total} & \textbf{212{,}561} & \textbf{2{,}132{,}648} & \textbf{40} & & & & & & & & \\
\bottomrule
\end{tabular}
}
\end{table*}

\subsection{Key Findings}

\begin{figure}[htbp]
    \centering
    \includegraphics[width=0.7\textwidth]{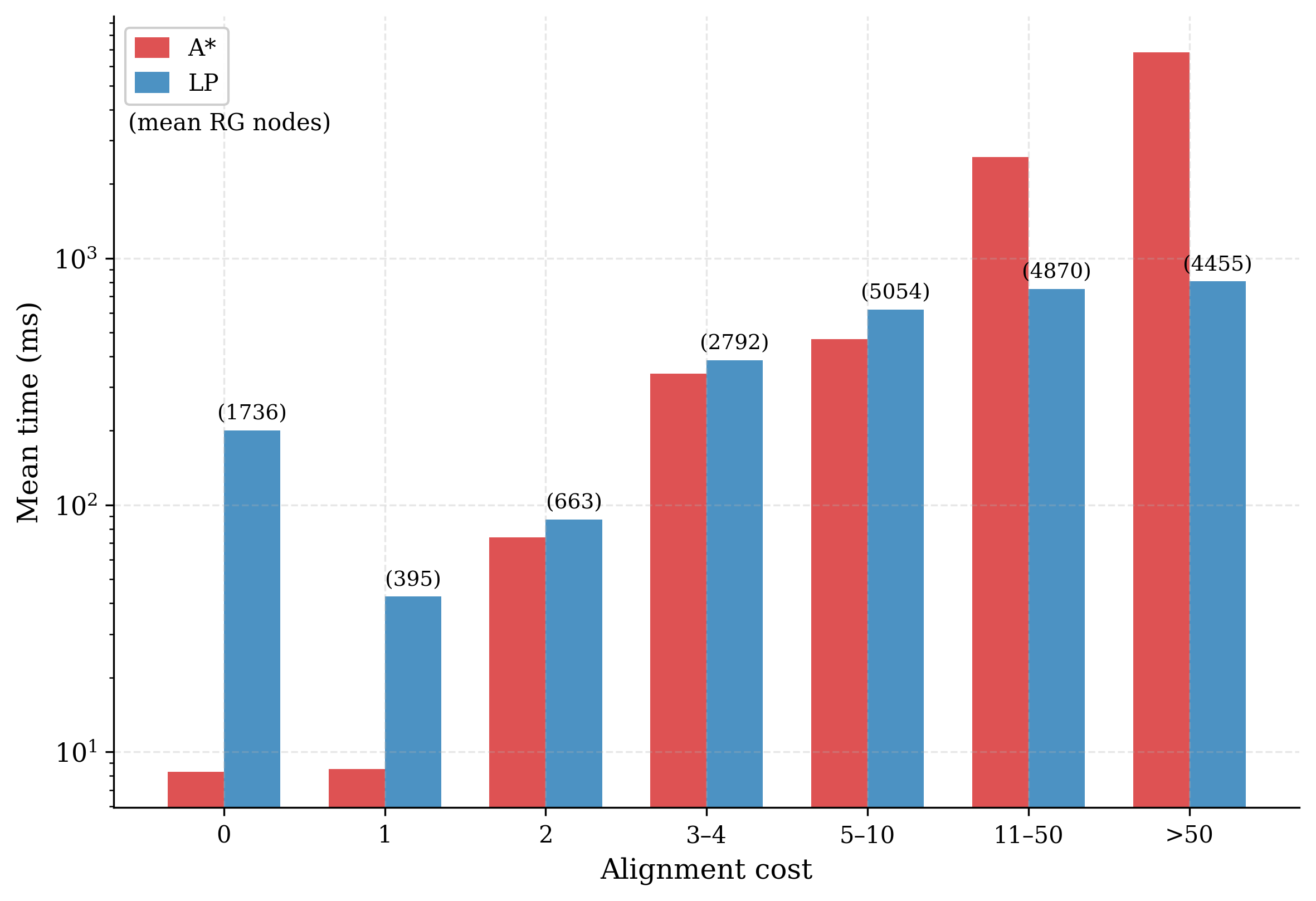}
    \caption{Mean execution time by alignment cost. Values in parentheses indicate mean RG size. LP time correlates with RG size rather than alignment cost, while $A^*$ time grows with the number of deviations.}
    \label{fig:time_by_alignment_cost}
\end{figure}

Our experiments reveal a hierarchy of factors that determine the relative performance of $A^*$ versus LP:
\begin{enumerate}
    \item Alignment cost (deviations): The dominant factor. When traces deviate significantly from the model (higher alignment costs), $A^*$'s search space explodes while LP maintains stable performance. Figure~\ref{fig:time_by_alignment_cost} illustrates this contrast: $A^*$ time grows exponentially with alignment cost, while LP time is governed by reachability graph size regardless of conformance level. This explains why LP is slower for perfectly conforming traces (cost 0) than for traces with 1--2 deviations. Cost-0 traces come from longer traces that produce larger reachability graphs (1,736 vs.\ 395 nodes). The effect of increasing alignment costs is dramatic, with $A^*$ slowing down up to 366$\times$ on noisy synthetic traces while LP time remains essentially constant. BPI Challenge 2013 confirms this on real-world data, with LP winning 96.7\% of traces with alignment cost 2--4.
    
    \item Model size (places): Strong effect on reachability graph size ($r = 0.79$). Small models (fewer than 15 places) favor LP regardless of trace length, as the reachability graph remains compact. For large models with many places and specifically many silent transitions, reachability graph construction effort becomes dominant, favoring $A^*$.
    
    \item Trace length: LP's advantage increases with trace length due to $A^*$'s exponential complexity in the worst case. A clear crossover point exists for datasets with sufficient trace length variation.
    
    \item Model precision: Moderate and model-dependent effect on reachability-graph size. In general, lower-precision models tend to produce larger reachability graphs because they permit more behavioral alternatives. However, this relationship is not monotone in our experiments, since precision co-varies with other structural properties such as silent transitions and model size. Accordingly, precision has only a weak direct effect on LP win rate (\(r=0.07\)) and fitness/alignment cost remains the main factor.
    
\end{enumerate}

\subsection{Per-Dataset Analysis}

\subsubsection{BPI Challenge 2012 (Loan Application)}

BPI Challenge 2012 demonstrates a crossover behavior. With 13{,}087 traces evaluated against 4 model variants, LP becomes increasingly competitive as trace length grows. The models range from 25--39 places with fitness 0.81--0.95 and precision 0.66--0.75.

\begin{figure}[ht]
    \centering
    \includegraphics[width=0.7\textwidth]{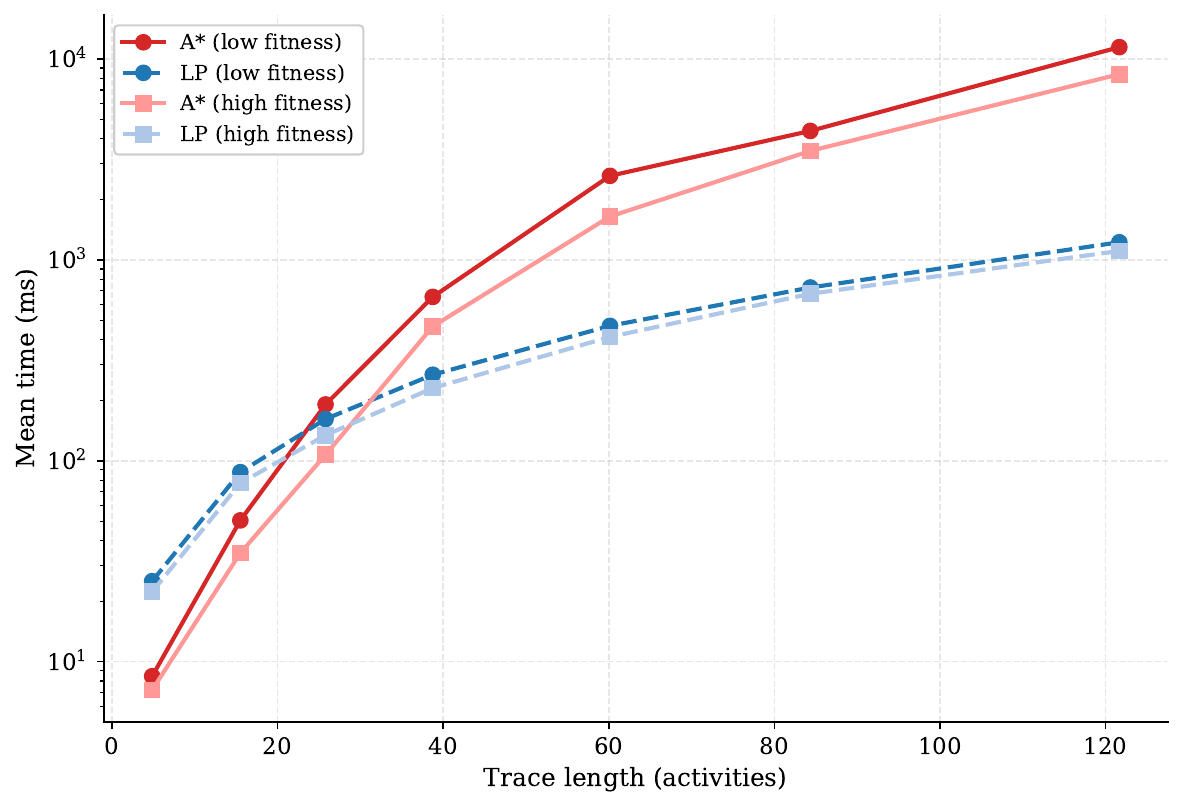}
    \caption{Mean execution time vs.\ trace length for $A^*$ and LP on 
    BPI Challenge 2012, stratified by model fitness. Both algorithms 
    exhibit polynomial scaling, but $A^*$ grows faster 
    than LP, 
    with crossover occurring at approximately 25--30 activities. Fitness threshold: 0.9.}
    \label{fig:bpi2012_time_vs_length}
\end{figure}

Table~\ref{tab:bpi2012-crossover} presents LP win rate, speedup, and detailed time breakdown by trace length. Figure~\ref{fig:bpi2012_time_vs_length} illustrates the scaling behavior of $A^*$ and LP as trace length increases.

\begin{table}[htbp]
\centering
\small
\setlength{\tabcolsep}{4pt}
\caption{BPI Challenge 2012: LP win rate, speedup of LP with respect to $A^*$, and time breakdown by trace length. RG  = reachability graph construction time; Solve = LP solver time.}
\label{tab:bpi2012-crossover}
\resizebox{0.95\textwidth}{!}{%
\begin{tabular}{@{}lrrrrrrrr@{}}
\toprule
\makecell[l]{\textbf{Trace}\\\textbf{Length}} & \textbf{Instances} & \makecell[r]{\textbf{LP Win}\\\textbf{(\%)}} & \textbf{Speedup} & \makecell[r]{\textbf{$A^*$}\\\textbf{(ms)}} & \makecell[r]{\textbf{LP Total}\\\textbf{(ms)}} & \makecell[r]{\textbf{RG}\\\textbf{(ms)}} & \makecell[r]{\textbf{Solve}\\\textbf{(ms)}} & \makecell[r]{\textbf{RG}\\\textbf{Nodes}} \\
\midrule
1--10 & 25{,}868 & 0.1 & 0.33$\times$ & 7.8 & 23.7 & 9.1 & 14.4 & 307 \\
11--20 & 5{,}404 & 6.0 & 0.52$\times$ & 42.6 & 82.5 & 30.9 & 51.3 & 867 \\
21--30 & 7{,}388 & 43.2 & 1.01$\times$ & 148.8 & 147.5 & 58.0 & 89.0 & 1{,}407 \\
31--50 & 9{,}632 & 67.5 & 2.25$\times$ & 559.4 & 248.8 & 105.0 & 143.2 & 2{,}087 \\
51--100 & 3{,}820 & 89.1 & 5.03$\times$ & 2{,}456 & 488.4 & 214.8 & 272.8 & 3{,}442 \\
$>$100 & 236 & 99.2 & 8.51$\times$ & 9{,}903 & 1{,}164 & 577.2 & 585.2 & 6{,}440 \\
\bottomrule
\end{tabular}
}
\end{table}

The crossover point occurs at approximately 25 activities, where LP begins to outperform $A^*$ in more than 50\% of cases. For traces exceeding 100 activities, LP achieves a 99.2\% win rate with an average speedup of 8.51$\times$ over $A^*$. This confirms the theoretical advantage of LP's polynomial complexity over $A^*$'s exponential worst-case behavior. The time breakdown reveals that reachability graph construction accounts for approximately 40--50\% of LP time, providing additional speedup opportunities.

\subsubsection{BPI Challenge 2013 (Incident Management)}

BPI Challenge 2013 provides a unique complement to BPI 2012, demonstrating that alignment cost, in addition to trace length, is a dominant factor determining LP performance. With 7{,}554 unique traces evaluated against 10 model variants (75{,}540 instances), LP achieves a 29.5\% overall win rate despite short average trace lengths (8.7 activities).

Table~\ref{tab:bpi2013-cost} presents LP performance segmented by alignment cost. The results reveal a striking pattern: LP wins only 2.6\% of traces with zero alignment cost (perfect fitness), but 96.7\% win rate on traces with alignment cost 2--4.

\begin{table}[htbp]
\centering
\small
\setlength{\tabcolsep}{6pt}
\caption{BPI Challenge 2013: LP win rate by alignment cost, demonstrating the dominant role of trace-model deviations. Speedup refers to $A^*$ with respect to LP times.}
\label{tab:bpi2013-cost}
\begin{tabular}{@{}lrrr@{}}
\toprule
\textbf{Alignment Cost} & \textbf{Instances} & \textbf{LP Win (\%)} & \textbf{Speedup} \\
\midrule
0 (perfect) & 44{,}631 & 2.6 & 0.32$\times$ \\
1 & 9{,}677 & 32.9 & 0.92$\times$ \\
2--4 & 16{,}657 & 96.7 & 1.63$\times$ \\
5--10 & 2{,}931 & 46.8 & 1.43$\times$ \\
$>$10 & 1{,}644 & 25.7 & 0.87$\times$ \\
\bottomrule
\end{tabular}
\end{table}

We wondered why LP win rate drops for high costs ($\ge 5$). Analysis reveals that traces with cost $>$10 come almost exclusively from the 2-place model (fitness 0.308), which has very small reachability graphs (mean 29 nodes). With such a compact model, $A^*$ remains efficient even with high alignment costs because the branching factor is minimal. 
The drop in LP win rate for costs 5--10 reflects a similar phenomenon: this range contains a mixture of traces from both compact models (where $A^*$ excels regardless of cost) and larger models (where LP maintains its advantage). Specifically, the 2-place and 8-place models contribute to this cost range, diluting LP's overall win rate. In contrast, costs 2--4 are dominated by traces from the 12-place model (fitness 0.902, mean 135 RG nodes), where LP achieves 98.2\% win rate (see, Table~\ref{tab:bpi2013-models}). This confirms that model size modulates the alignment-cost effect: LP's advantage on nonconforming traces is strongest when reachability graphs are sufficiently large to offset construction overhead.

Table~\ref{tab:bpi2013-models} provides a thorough analysis across different model configurations, revealing how fitness, precision, and size interact. For example, the 8-place models provide a clear demonstration: all have identical RG size (77 nodes) but vary dramatically in LP win rate based solely on fitness. The model with fitness 0.940 (average alignment cost was 1.8) achieves 99.4\% LP win rate, while the model with fitness 0.999 (average alignment cost was 0.0) achieves only 6.8\%. This 15$\times$ difference in LP win rate is entirely explained by alignment cost.

\begin{table}[htbp]
\centering
\small
\setlength{\tabcolsep}{5pt}
\caption{BPI Challenge 2013: Model characteristics and algorithm performance. Precision affects RG size; fitness (via alignment cost) determines LP win rate.}
\label{tab:bpi2013-models}
\begin{tabular}{@{}rrrrrr@{}}
\toprule
\textbf{Places} & \textbf{Fitness} & \textbf{Precision} & \textbf{RG Nodes} & \textbf{Mean Cost} & \textbf{LP Win (\%)} \\
\midrule
2 & 0.308 & 0.600 & 29 & 7.7 & 57.9 \\
8 & 0.940 & 0.924 & 77 & 1.8 & 99.4 \\
8 & 0.997 & 0.912 & 77 & 0.1 & 12.9 \\
8 & 0.997 & 0.885 & 77 & 0.1 & 11.4 \\
8 & 0.999 & 0.883 & 77 & 0.0 & 6.8 \\
10 & 0.925 & 0.831 & 174 & 0.4 & 0.1 \\
10 & 0.929 & 0.716 & 116 & 0.5 & 4.0 \\
11 & 0.996 & 0.624 & 135 & 0.1 & 3.3 \\
12 & 0.902 & 0.723 & 135 & 2.2 & 98.2 \\
17 & 1.000 & 0.626 & 871 & 0.0 & 0.6 \\
\bottomrule
\end{tabular}
\end{table}

\subsubsection{Sepsis Cases (Hospital Emergency Care)}

The Sepsis dataset (1{,}050 unique traces $\times$ 9 model variants = 9{,}450 instances, 9{,}010 valid) presents a challenging case for LP due to large reachability graphs (mean 8{,}215 nodes). The models range from 22--34 places with fitness 0.88--0.93 and precision 0.52--0.70. Despite high LP overhead, LP achieves a 19.1\% overall win rate, with performance improving significantly for longer traces.

Table~\ref{tab:sepsis-crossover} presents LP performance stratified by trace length.

\begin{table}[htbp]
\centering
\setlength{\tabcolsep}{5pt}
\caption{Sepsis Cases: LP win rate by trace length, showing crossover at approximately 20--25 activities. Speedup refers to $A^*$ with respect to LP times.}
\label{tab:sepsis-crossover}
\resizebox{0.95\textwidth}{!}{%
\begin{tabular}{@{}lrrrrrr@{}}
\toprule
\textbf{Trace Length} & \textbf{Instances} & \textbf{LP Win (\%)} & \textbf{Speedup} & \textbf{Mean $A^*$} & \textbf{Mean LP} \\
\midrule
1--20 & 7{,}909 & 16.2 & 0.34$\times$ & 350\,ms & 1{,}045\,ms \\
21--50 & 1{,}040 & 39.1 & 1.34$\times$ & 2{,}679\,ms & 1{,}993\,ms \\
$>$50 & 61 & 60.7 & 6.98$\times$ & 25.3\,s & 3{,}631\,ms \\
\bottomrule
\end{tabular}
}
\end{table}

The crossover occurs at approximately 20--25 activities, higher than BPI 2012 due to the larger reachability graphs. For the longest traces ($>$50 activities), LP achieves nearly 7$\times$ speedup.
Note that 95.3\% of instances produced optimal alignments from both methods; the remaining 4.7\% timed out during LP reachability graph construction on the most complex model variant (34 places).

\subsubsection{Road Traffic Fine Management}

Road Traffic Fine, which includes very short traces (mean 3.7 activities, max 20), was evaluated using 13 model variants. Combined with high fitness models, this creates conditions strongly favoring $A^*$. LP win rate, which is only 2.7\%, is associated with the longer traces (15-20 activities) from lower-fitness model variants where alignment costs exceed 2-3 deviations.

\subsubsection{PLG2 Synthetic Benchmark}
We mainly selected PLG2 synthetic datasets to provide controlled experiments that highlight the role of alignment cost, and fitness as its proxy, in determining algorithm performance.
The benchmark comprises 21 log files based on 4 unique process models, with clean and noisy trace variants at different trace lengths. Each log is evaluated against its reference model (hence instances = traces = 40{,}500). Of these, 40{,}230 instances (99.3\%) produced optimal alignments from both methods.

From this benchmark, Table~\ref{tab:clean-vs-noisy} compares clean (perfect fitness) and noisy variants across representative models.

\begin{table*}[htbp]
\centering
\setlength{\tabcolsep}{4pt}
\caption{PLG2 synthetic benchmark: clean vs.\ noisy trace variants. Each row pair shows the same model with clean traces (perfect fitness) and noisy traces (with injected deviations). $A^*$ slowdown indicates the ratio of noisy to clean $A^*$ runtime.}
\label{tab:clean-vs-noisy}
\resizebox{\textwidth}{!}{%
\begin{tabular}{@{}llrrrrrrrc@{}}
\toprule
\textbf{Model} & \textbf{Variant} & \textbf{Traces} & \textbf{Places} & \textbf{Fitness} & \makecell[r]{\textbf{Mean}\\\textbf{length}} & \makecell[r]{\textbf{Mean}\\\textbf{$A^*$}} & \makecell[r]{\textbf{Mean}\\\textbf{LP}} & \makecell[r]{\textbf{LP win}\\\textbf{(\%)}} & \makecell[r]{\textbf{$A^*$}\\\textbf{slowdown}} \\
\midrule
A48-m37 & Clean & 2{,}000 & 55 & 1.000 & 36.5 & 13.4\,ms & 728.5\,ms & 0.2 & -- \\
 & Noisy & 2{,}000 & 55 & 0.982 & 36.2 & 685.0\,ms & 667.8\,ms & 19.1 & 51$\times$ \\
\midrule
A48-m50 & Clean & 2{,}000 & 55 & 1.000 & 49.8 & 14.4\,ms & 981.0\,ms & 0.1 & -- \\
 & Noisy & 2{,}000 & 55 & 0.982 & 49.9 & 2{,}745\,ms & 1{,}046\,ms & 34.8 & 190$\times$ \\
\midrule
A32-m27 & Clean & 2{,}000 & 37 & 1.000 & 26.7 & 7.4\,ms & 109.5\,ms & 0.4 & -- \\
 & Noisy & 2{,}000 & 37 & 0.966 & 26.7 & 75.4\,ms & 106.4\,ms & 31.6 & 10$\times$ \\
\midrule
A32-m41 & Clean & 2{,}000 & 37 & 1.000 & 41.3 & 9.8\,ms & 189.5\,ms & 0.1 & -- \\
 & Noisy & 2{,}000 & 37 & 0.956 & 40.9 & 589.4\,ms & 185.6\,ms & 63.9 & 60$\times$ \\
\midrule
A57-m52 & Clean & 2{,}000 & 64 & 1.000 & 52.3 & 15.4\,ms & 1{,}856\,ms & 0.0 & -- \\
 & Noisy & 2{,}000 & 64 & 0.971 & 51.1 & 5{,}635\,ms & 1{,}759\,ms & 41.2 & 366$\times$ \\
\bottomrule
\end{tabular}
}
\end{table*}

On clean traces (perfect fitness), $A^*$ dominates with sub-millisecond execution times because the heuristic guides search directly to the optimal alignment. However, when noise is introduced:

\begin{itemize}[label=--,nosep]
    \item $A^*$ slows down by 10--366$\times$ relative to its clean-trace performance on the same model due to increased search efforts,
    \item LP time remains essentially unchanged (within 10\%) because the reachability graph size is independent of alignment cost, and
    \item LP win rate increases from $<$1\% to 19--64\%.
\end{itemize}

Aggregating across all PLG2 logs: clean traces (11 logs, 20{,}230 instances) show 0.2\% LP win rate with mean $A^*$ time of 32.2\,ms, while noisy traces (10 logs, 20{,}000 instances) show 28.1\% LP win rate with mean $A^*$ time of 1{,}234\,ms---a 38$\times$ average slowdown for $A^*$.

\subsection{LP-Dominant Conditions}

Based on our experimental analysis, Table~\ref{tab:lp-dominant} identifies the conditions under which LP consistently outperforms $A^*$.

\begin{table}[htbp]
\centering
\small
\setlength{\tabcolsep}{5pt}
\caption{LP-dominant subsets: Performance when restricting to favorable conditions. Speedup refers to $A^*$ with respect to LP times.}
\label{tab:lp-dominant}
\begin{tabular}{@{}llrrr@{}}
\toprule
\textbf{Dataset} & \textbf{Condition} & \textbf{Instances} & \textbf{LP Win (\%)} & \textbf{Speedup} \\
\midrule
BPI Challenge 2012 & Trace $>$ 50 & 4{,}056 & 89.7 & 5.47$\times$ \\
BPI Challenge 2012 & Trace $>$ 100 & 236 & 99.2 & 8.51$\times$ \\
BPI Challenge 2013 & Cost $\geq$ 2 & 21{,}232 & 84.3 & 1.58$\times$ \\
Sepsis Cases & Trace $>$ 50 & 61 & 60.7 & 6.98$\times$ \\
PLG2 A48-m50-noise & Trace $>$ 50 & 878 & 79.4 & 3.33$\times$ \\
\bottomrule
\end{tabular}
\end{table}

To understand LP performance bottlenecks, we analyzed the time distribution between reachability graph construction and LP solving. Table~\ref{tab:lp-breakdown} includes $A^*$ time for direct comparison.

\begin{table}[ht]
\centering
\small
\setlength{\tabcolsep}{5pt}
\caption{LP time breakdown: RG construction vs.\ LP solving, with $A^*$ comparison. RG (\%) is computed as RG Build / (RG Build + LP Solve). remaining time ($\sim$5--10\%) includes synchronous product construction and result extraction overhead.}
\label{tab:lp-breakdown}
\begin{tabular}{@{}lrrrrr@{}}
\toprule
\textbf{Dataset} & \makecell[r]{\textbf{$A^*$}\\\textbf{(ms)}} & \makecell[r]{\textbf{RG Build}\\\textbf{(ms)}} & \makecell[r]{\textbf{LP Solve}\\\textbf{(ms)}} & \makecell[r]{\textbf{RG}\\\textbf{(\%)}} & \makecell[r]{\textbf{RG}\\\textbf{Nodes}} \\
\midrule
Sepsis Cases & 788  & 359 & 813  & 30.6 & 8{,}215 \\
BPI Challenge 2012 & 356 & 54 & 74 & 42.0 & 1{,}104 \\
BPI Challenge 2013 & 8.6 & 5.8  & 12.1 & 32.1 & 177 \\
Road Traffic Fine & 9.3 & 37.0 & 76.2 & 32.7 & 797 \\
\bottomrule
\end{tabular}
\end{table}

\begin{figure}[htbp]
    \centering
    \includegraphics[width=0.75\textwidth]{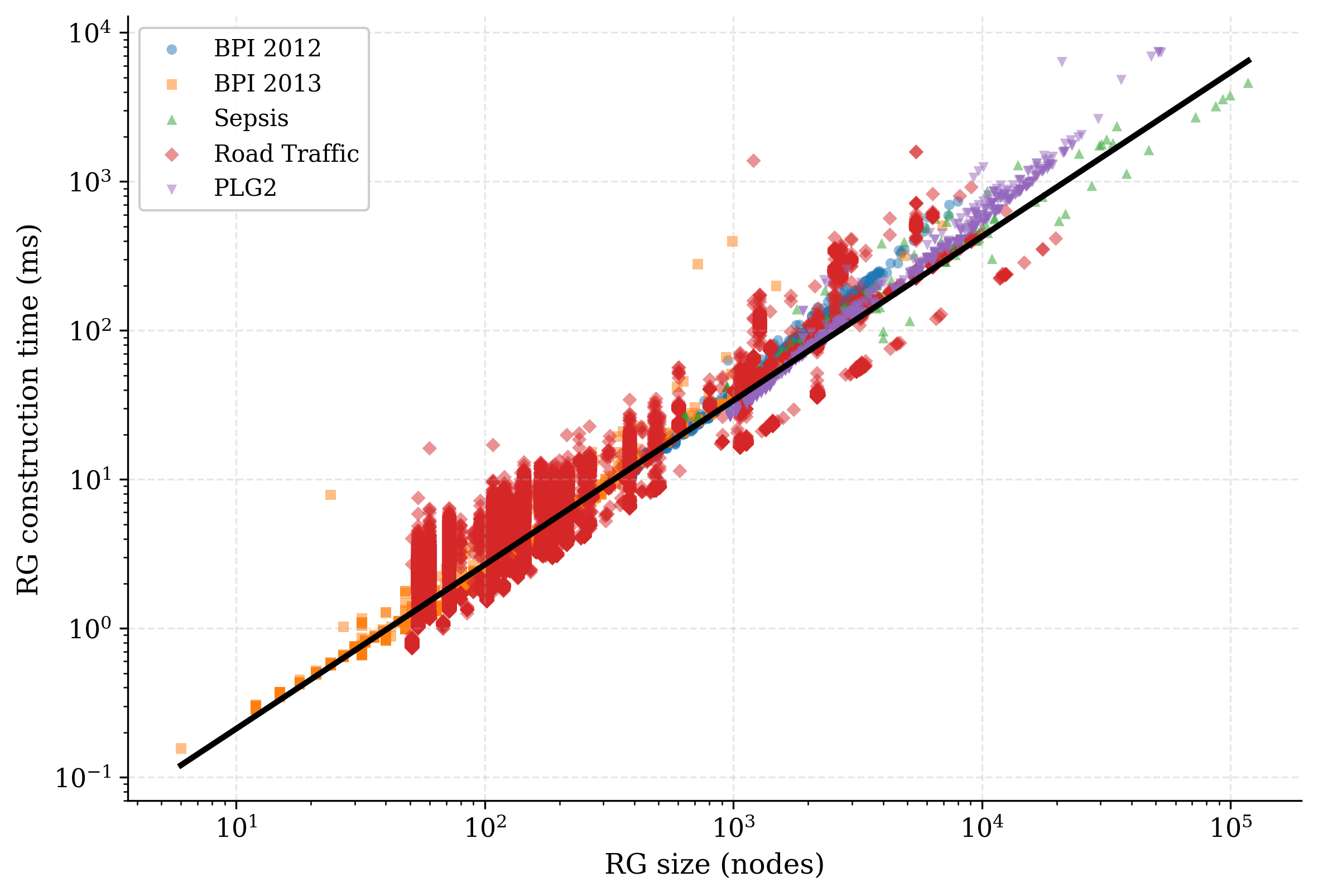}
    \caption{RG construction time vs.\ graph size across all datasets. The fitted line shows near-linear scaling ($R^2 = 0.93$).}
    \label{fig:rg_time_vs_size}
\end{figure}

Generally, reachability graph construction accounts for more than 30\%, and sometimes much more, of total LP time, which includes synchronous product construction, reachability graph enumeration, LP solving, and result extraction overhead. Figure~\ref{fig:rg_time_vs_size} shows that RG construction time scales near-linearly with graph size across all datasets, confirming predictable overhead growth. The size of the reachability graph, measured as the number of reachable markings (states), is jointly determined by process model size and trace length. A partial correlation analysis across all datasets, correlating RG node count with process model places (controlling for trace length) and with trace length (controlling for model places), reveals that the dominant factor varies by dataset characteristics: 
in datasets with high trace length variability (e.g., BPI 2012 with traces ranging 3--175 activities), trace length explains nearly all variance in RG size ($R^2 = 0.99$); in datasets with constrained trace lengths (e.g., Road Traffic Fine), model size (number of places) becomes more influential ($R^2 = 0.50$ vs $0.05$ for trace length).
This dataset-dependent relationship explains why LP overhead is governed primarily by model size for short-trace datasets, while trace length becomes the critical bottleneck for datasets with longer, more variable traces.

\subsection{Summary of Findings}

The key findings are:

\begin{itemize}[label=--,nosep]
    \item Correctness: For all instances solved by both methods, the LP objective value matched the $A^*$ objective value exactly, providing strong empirical evidence that the LP formulation returns the same optimal alignment cost as $A^*$ on the benchmarked solved cases.
    
    \item Scalability: LP exhibits polynomial complexity, achieving up to 8.5$\times$ speedup over $A^*$ on long traces.
    
    \item Robustness to deviations: Unlike $A^*$, LP performance is largely independent of alignment cost. On noisy traces where $A^*$ slows down 10--366$\times$, LP time remains constant. BPI 2013 confirms this finding on real-world data: LP achieves 96.7\% win rate on traces with alignment cost 2--4.

    \item Complementarity: $A^*$ and LP exhibit complementary performance 
    characteristics, with $A^*$ excelling on short, conforming traces and LP on longer traces with deviations. This motivates a selection formula developed in Section~\ref{sec:algorithm-selection}.

\end{itemize}

The key practical insight is that LP excels in situations most relevant to conformance checking: long traces with significant deviations from the process model.

\section{Algorithm Selection}
\label{sec:algorithm-selection}

We propose a selection formula that predicts the faster method using features available before execution. The formula builds on two observations: (1) $A^*$ exhibits exponential degradation with increasing deviations while LP maintains stable performance, and (2) LP's reachability graph construction overhead pays off only for sufficiently long traces.

Since alignment cost is unknown before execution, we use model fitness $F$ as a proxy. The expected number of deviations for a trace of length $L$ can be approximated as $(1-F) \times L$, representing the proportion of non-conformant activities multiplied by trace length. This yields the following selection rule:
\begin{equation}
\text{Use LP when: } L > 20 \land (1-F) \times L > 1.5
\label{eq:selection-formula}
\end{equation}
\noindent where $L$ is the trace length and $F$ is the model fitness. The first condition ensures traces are long enough to justify LP's construction overhead, while the second targets traces where $A^*$ is expected to struggle due to high alignment costs. 

The proposed selection rule is not intended as a universal threshold that is
optimal across all datasets, model-discovery pipelines, and cost settings.
Rather, it should be viewed as an empirically grounded practical guideline
whose cross-dataset robustness is confirmed by the leave-one-dataset-out
analysis reported below.

\begin{figure}[htbp]
    \centering
    \includegraphics[width=0.75\textwidth]{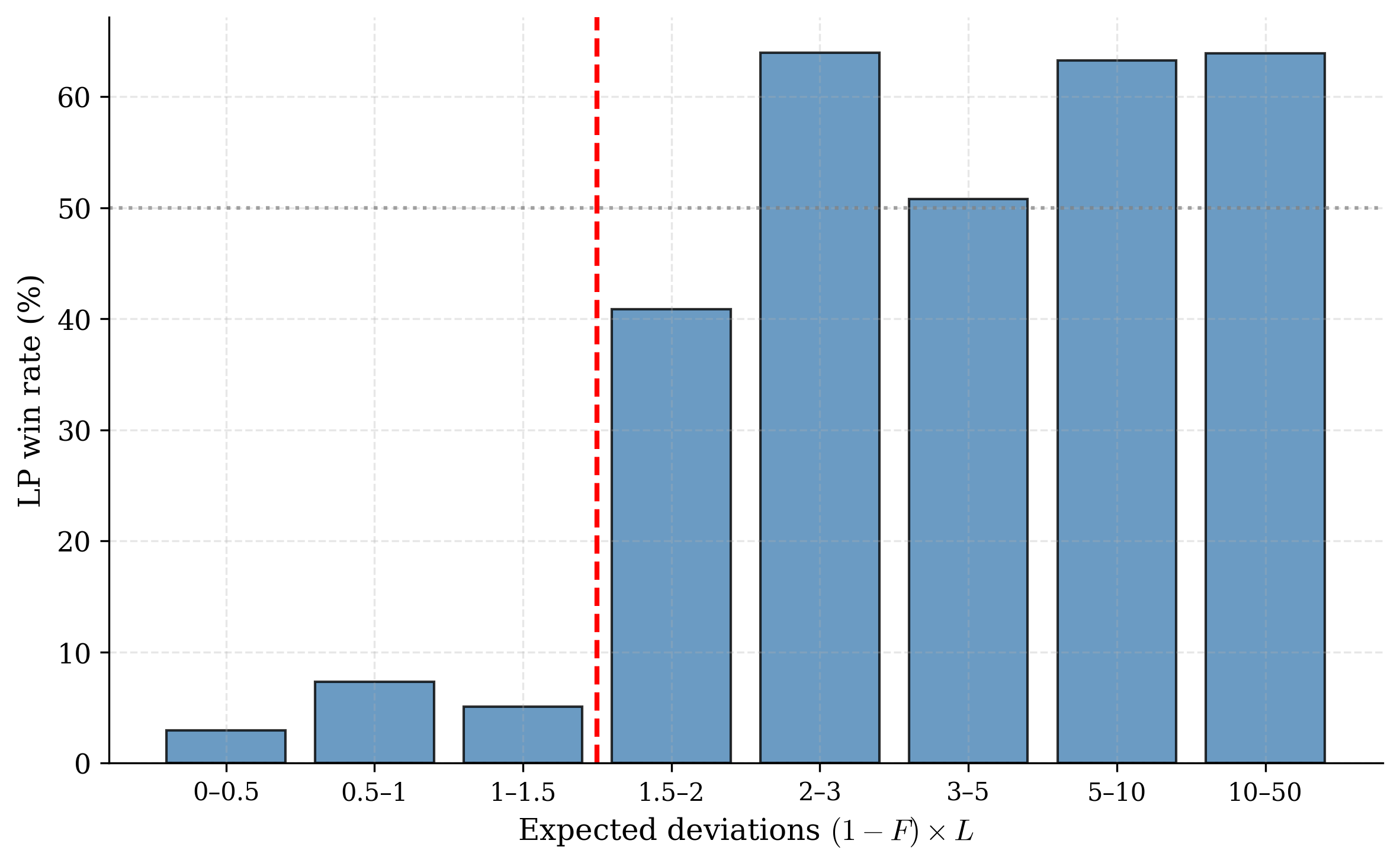}
   \caption{LP win rate by expected deviations across all datasets. 
    The dashed line marks the selection threshold at $(1-F) \times L = 1.5$.}
    \label{fig:lp_winrate_expected_dev}
\end{figure}

Figure~\ref{fig:lp_winrate_expected_dev} validates this threshold: LP win rate
jumps from 3--7\% below the threshold to 41--64\% above it.

The thresholds in Equation~\eqref{eq:selection-formula} were obtained through a grid search over
$L_{\mathrm{thresh}} \in \{15, 20, 25, 30, 35, 40\}$ and
$\mathrm{dev}_{\mathrm{thresh}} \in \{0.5, 1.0, 1.5, 2.0, 2.5\}$, selecting the
combination that maximized total time savings relative to always using~$A^*$
while avoiding regression on any individual dataset. The selected combination
lies in a stable region of the parameter space: nearby alternatives produce
comparable savings (Table~\ref{tab:formula-comparison}), indicating that the
result is not sensitive to the precise threshold values.
To verify that this stability extends across datasets, we performed
leave-one-dataset-out (LODO) cross-validation: in each of five folds the grid
search was re-run on four datasets and the winning thresholds were evaluated on
the held-out dataset. Every fold selected $L \in \{20, 25\}$ and
$\mathrm{dev} \in \{1.0, 1.5\}$, and
Equation~\eqref{eq:selection-formula} matched or outperformed the LODO-optimal
rule on four of five held-out datasets. The single exception (BPI~2013)
differed by only 0.5~percentage points. On the Sepsis fold, where the
LODO-optimal rule chose a more aggressive deviation threshold ($1.0$ vs.\
$1.5$), Equation~\eqref{eq:selection-formula} achieved higher savings
($+28.4\%$ vs.\ $+20.0\%$) because the conservative threshold avoids
selecting LP on traces with prohibitively large reachability graphs.
We note that while model precision and silent transitions influence LP overhead,
they did not improve the selection rule consistently beyond trace length and
fitness in our experiments. We therefore retain the two-feature rule for
interpretability; richer selection models incorporating structural features are
left as future work.

Consequently, LP is not selected for long traces from high-fitness models (e.g., $L=100$ with $F=0.99$ yields $(1-F) \times L = 1.0 < 1.5$), where $A^*$ remains efficient despite trace length because few deviations are expected.

\subsection{Empirical Validation}

We evaluated the selection formula across the four real-world datasets by simulating a hybrid approach: for each trace, we use the time of whichever algorithm the formula selects. Table~\ref{tab:selection-results} presents the results compared to always using $A^*$ (the conventional approach) and the theoretical oracle that always selects the faster algorithm.

\begin{table}[htbp]
\centering
\small
\setlength{\tabcolsep}{5pt}
\caption{Algorithm selection performance across datasets. Savings are relative to always using $A^*$. The hybrid approach applies Equation~\ref{eq:selection-formula}.}
\label{tab:selection-results}
\begin{tabular}{@{}lrrrrr@{}}
\toprule
\textbf{Dataset} & \textbf{Instances} & \textbf{Always $A^*$} & \textbf{Hybrid} & \textbf{Oracle} & \textbf{Savings} \\
\midrule
Sepsis Cases & 9{,}010 & 7{,}101s & 5{,}082s & 3{,}619s & +28.4\% \\
BPI Challenge 2012 & 52{,}348 & 18{,}637s & 5{,}952s & 5{,}153s & +68.1\% \\
BPI Challenge 2013 & 75{,}540 & 650s & 658s & 533s & $-$1.2\% \\
Road Traffic Fine & 1{,}954{,}810 & 11{,}665s & 11{,}665s & 9{,}079s & +0.0\% \\
\midrule
\textbf{Total} & \textbf{2{,}091{,}708} & \textbf{38{,}053s} & \textbf{23{,}356s} & \textbf{18{,}384s} & \textbf{+38.6\%} \\
\bottomrule
\end{tabular}
\end{table}

The formula achieves 38.6\% time savings overall, with substantial gains on datasets where LP has clear advantages (BPI 2012: +68.1\%, Sepsis: +28.4\%) while avoiding major slowdowns. The formula shows a small slowdown on BPI 2013 ($-$1.2\%) due to the dataset being dominated by short, high-fitness traces where $A^*$ is efficient.
The formula never selects LP for Road Traffic Fine traces because the combination of short traces (mean $L=3.7$) and fitness ($F \geq 0.77$) keeps $(1-F) \times L$ below the threshold.

We compared the proposed formula against simpler alternatives in Table~\ref{tab:formula-comparison}. The fitness-based formula outperforms pure length thresholds by correctly identifying when $A^*$ will struggle regardless of trace length.

\begin{table}[htbp]
\centering
\small
\setlength{\tabcolsep}{5pt}
\caption{Comparison of selection formulas. Per-dataset savings shown as percentage improvement over always using $A^*$.}
\label{tab:formula-comparison}
\begin{tabular}{@{}lrrrrr@{}}
\toprule
\textbf{Formula} & \textbf{Sepsis} & \textbf{BPI 2012} & \textbf{BPI 2013} & \textbf{Road} & \textbf{Total} \\
\midrule
$L > 30$ & +20.0 & +67.4 & $-$15.0 & +0.0 & +36.5 \\
$(1-F) \times L > 2$ & +19.6 & +66.4 & $-$0.5 & $-$0.1 & +36.1 \\
Equation~\ref{eq:selection-formula} & +28.4 & +68.1 & $-$1.2 & +0.0 & \textbf{+38.6} \\
\bottomrule
\end{tabular}
\end{table}

Notably, the simple threshold $L > 30$ causes a 15.0\% slowdown on BPI 2013 because it triggers LP on long traces from high-fitness models where $A^*$ is faster. The fitness-based formula substantially reduces this slowdown by recognizing that high fitness implies low alignment costs. Nevertheless, using even simple rules for selecting between $A^*$ and LP can lead to substantial time savings.

\subsection{Limitations and Error Analysis}

Compared to an oracle who always selects the right option, the formula achieves 96.0\% classification accuracy, with two types of errors:

\paragraph{False Positives (0.36\% of the instances)} The formula selects LP but $A^*$ was faster. These 7{,}557 instances occur predominantly on low-precision models ($< 0.6$), where the reachability graph becomes large despite high fitness. In such cases, LP's graph construction overhead exceeds $A^*$'s search time. The total time cost is 1{,}649 seconds with median cost of 92\,ms per error.

\paragraph{False Negatives (3.64\% of instances)} The formula selects $A^*$ but LP was faster. These 76{,}181 instances occur almost exclusively on short traces ($L \leq 20$), which are filtered by the length threshold. Crucially, 94.3\% of false negatives have costs below 10\,ms, meaning that the formula correctly identifies that LP's overhead is not justified for such traces even when LP would technically be faster. The total time cost is 3{,}323 seconds with median cost of only 0.7\,ms per error.

\paragraph{Safe Mistakes} The formula prioritizes avoiding high LP overhead over capturing every LP opportunity. Using LP on high-fitness, low-precision models can result in significant slowdowns, as LP must construct enormous reachability graphs for models that $A^*$ handles efficiently. The formula's conservative thresholds sacrifice potential gains on short traces (where costs are typically small) to prevent these high losses.

The formula produces a net benefit of 38.6\% improvement in the total computation time compared to always using $A^*$. The total time cost of formula errors is 4{,}972 seconds, compared to the unattained oracle's correct selections. 

\subsection{Practical Considerations}

The selection formula requires only trace length and model fitness, both
available before conformance checking. Model fitness is typically computed
during process discovery or can be estimated efficiently using token-based
replay. The formula adds negligible computational overhead while providing
substantial time savings on datasets where LP excels.

For practitioners, we recommend the formula as a default selection strategy
when both algorithms are available. On datasets dominated by short, conformant
traces, the formula will conservatively default to~$A^*$, incurring no penalty.
On datasets with longer traces or lower fitness, it will identify opportunities
for LP acceleration. The formula can also be tuned: lowering the length
threshold to $L > 15$ captures more LP opportunities at the cost of increased
false positives, while raising the deviation threshold to $(1-F) \times L > 2$
provides more conservative LP selection. The LODO cross-validation reported
above confirms that the selected thresholds are not dataset-specific: all five
held-out folds recovered thresholds within the same narrow region, supporting
the use of Equation~\eqref{eq:selection-formula} as a default starting point
for new datasets and model configurations. Future work may explore richer
selection models that incorporate additional structural features, such as
precision, silent transitions, or other proxies for reachability-graph growth.
In this paper, we deliberately retain a simple and interpretable rule based on
features that are readily available before conformance checking.

\section{Limitations and Future Directions}
\label{sec:limitations}


While our experiments demonstrate the complementary strengths of LP and $A^*$ for conformance checking, both approaches exhibit limitations that warrant discussion.

We deliberately compare \model{} against standard $A^*$ as implemented in \texttt{PM4Py}, the practical baseline most practitioners encounter, rather than specialized variants. Our central insight that the LP defined over the reachability graph is solvable in polynomial time due to total unimodularity holds independently of the chosen baseline.

\subsection{Limitations of the LP Approach}

The LP approach's primary bottleneck is reachability graph construction, which consumed 30--42\% of total LP time across our experiments (Table~\ref{tab:lp-breakdown}). This overhead becomes prohibitive under two conditions:

\paragraph{Low-Precision Models} Models with low precision permit many behavioral paths not observed in the event log, resulting in large reachability graphs. In our experiments, models with precision below 0.6 produced reachability graphs where LP times can exceed $A^*$ by factors of 40 or more, even for traces where LP would theoretically find alignments faster. The selection formula mitigates this by avoiding LP on high-fitness models (where $A^*$ is efficient regardless of precision), but cannot fully address cases where LP would otherwise be advantageous.

\paragraph{Silent Transitions} Models with numerous silent ($\tau$) transitions exhibit combinatorial growth in reachability graph size, as each silent transition introduces additional state-space branches. This effect compounds with low precision, as silent transitions often appear in ``flower'' constructs that generalize model behavior. In the Sepsis dataset, models produced reachability graphs averaging 8{,}215 nodes, compared to 177 nodes for the simpler BPI 2013 models.

\paragraph{Memory Considerations} 
The LP approach stores the reachability graph in memory, unlike $A^*$'s incremental exploration. While memory did not emerge as a bottleneck in our experiments, this trade-off warrants future investigation.

\paragraph{Solver Selection.}
URC$^2$ uses the Gurobi solver, whereas \texttt{PM4Py}'s A$^*$ relies on cvxopt\_solver\_custom\_align, a lightweight backend specifically optimized for alignment computation.
This asymmetry reflects the differing computational profiles of the two approaches.
A$^*$ typically solves hundreds of small LPs per trace, where solvers with low per-call overhead are most efficient.
URC$^2$ solves a single, much larger LP over the reachability graph, where solver efficiency at scale matters.
In our experiments, equipping A$^*$ with Gurobi increased its runtime by a factor of more than 3 due to per-call overhead, confirming that the default configuration is already favorable to A$^*$.

More broadly, achieving a perfectly symmetric comparison is inherently
difficult. \texttt{PM4Py}'s $A^*$ benefits from CPython-compiled components (heap
operations, linear algebra) and years of engineering optimization. URC$^2$
accelerates the core successor-computation loops of reachability graph
construction via Numba JIT compilation, but the surrounding orchestration remains in Python. Overall, \texttt{PM4Py} represents a more mature and optimized implementation, and the reported comparison therefore favors~$A^*$ rather than~URC$^2$. Engineering differences
may shift the crossover point identified earlier, but the underlying
complementarity between $A^*$ and URC$^2$---driven by the exponential
sensitivity of $A^*$ to alignment cost versus the stability of the LP
approach---remains independent of implementation choices.

\paragraph{Control-Flow Conformance}
Our formulation addresses control-flow conformance only. The total unimodularity result relies on the reduction to a pure node--arc flow model on the reachability graph. In multi-perspective conformance checking, additional data, resources, or time constraints would generally appear as side constraints beyond standard flow conservation, which may destroy the network-flow structure and the corresponding TU guarantee. Extending the approach to such settings is therefore nontrivial and left for future work.

These limitations suggest several directions for future work, beyond the scope of this paper. Symbolic representations using Decision Diagrams \citep[][]{thierry2015symbolic} could compress the reachability graph exponentially for models with regular structure. Lazy construction strategies could generate only selected portions of the reachability graph guided by the trace or by lower-bound information, thereby reducing the explicit breadth-first exploration that currently dominates LP preprocessing time in our implementation. Additionally, GPU-accelerated graph construction could further accelerate this preprocessing stage.

\newpage
\subsection{Limitations of the $A^*$ Approach}

The $A^*$ algorithm exhibits complementary weaknesses:

\paragraph{Exponential Degradation} $A^*$ search time grows exponentially with deviations. Each deviation expands the search frontier as demonstrated by 10--366$\times$ slowdowns on noisy traces (Table~\ref{tab:clean-vs-noisy}). This degradation is fundamental to $A^*$'s best-first search strategy and cannot be fully addressed through heuristic improvements.

\paragraph{Unpredictable Pathologies} We observed cases where $A^*$ exhibited extreme runtimes on short, seemingly simple traces. These pathological cases arise from interactions between the trace, model structure, and search heuristic that cannot be predicted from features like trace length or model fitness.

\vspace{10pt}

A natural concern is whether the proposed LP formulation offers a genuine
advantage over $A^*$ search, given that $A^*$ does not require explicit
construction of the reachability graph, whereas the LP is defined over
such a graph. Although the reachability graph of a Petri net may be
exponential in the worst case, this difficulty is inherent to exact
conformance checking and is not specific to the LP approach. In our current
implementation, we explicitly construct a bounded reachability subgraph of the
synchronous product by breadth-first exploration and then solve the LP on that
subgraph. A more selective construction of only the ultimately relevant states
is not part of the current implementation and remains an interesting direction
for future work.

The essential difference therefore is not the elimination of state-space
explosion, but the optimization paradigm applied once a relevant portion of
the state space has been constructed. On the same constructed graph, $A^*$
still explores states through heuristic-guided branching, which may lead to
exponential runtimes under high branching factors or long optimal alignments.
In contrast, \model{} solves a single minimum-cost flow LP on the constructed
graph. Since the corresponding constraint matrix is totally unimodular, the LP
relaxation is integral, and the optimization problem is polynomially solvable
in the size of the constructed graph without integer variables. Hence, the
main contribution of the LP reformulation is not to avoid state-space
explosion, but to avoid additional combinatorial optimization during alignment
computation once the graph has been constructed. This distinction explains why
the LP formulation can outperform $A^*$ in practice on long, deviating traces.

LP and $A^*$ exhibit complementary performance characteristics that are predictable from model quality metrics, enabling an informed and systematic selection of the most suitable approach.

Our empirical evaluation benchmarks against \texttt{PM4Py}'s standard $A^*$ implementation. Advances within the $A^*$ paradigm---including the extended marking equation~\citep{vandongen2018efficiently}, cache-enhanced split-point search~\citep{li2022cache}, and REACH~\citep{casas2024reach}---improve heuristic quality and pruning, reducing search effort. A direct comparison with these
methods would be valuable but faces cross-framework challenges, as these techniques are implemented outside the Python ecosystem used by both \texttt{PM4Py} and URC$^2$. Such a comparison, alongside LP-side improvements such as model-level reachability graph caching and lazy graph construction, is a promising direction for future work. That said, we note that the complementarity between search-based and LP-based approaches is structural: improved $A^*$ variants shift the crossover point but do not eliminate the exponential sensitivity of best-first search to alignment cost.

\section{Conclusion}
\label{sec:conclusion}

This paper demonstrated that a totally unimodular LP reformulation enables
exact alignment computation on an explicitly constructed reachability graph
without integer constraints. In particular, once that graph has been
constructed, the resulting optimization subproblem is polynomially solvable.

Extensive empirical evaluation confirms that LP can significantly outperform classical $A^*$ search in regimes characterized by long traces and substantial deviations, achieving up to 8.5$\times$ speedup and 99\% win rates.

We developed an algorithm-selection formula based on trace length and model fitness that achieves 38.6\% time savings over always using $A^*$, with 96\% accuracy in identifying when LP's polynomial complexity advantage outweighs its construction overhead.

The results provide clear, data-driven guidance for off-the-shelf solver selection: use LP for long traces ($L > 20$) with expected deviations $(1-F) \times L > 1.5$, and default to $A^*$ otherwise. Leave-one-dataset-out cross-validation confirms that these thresholds
generalize across datasets with diverse characteristics, enabling practitioners
to adopt the rule as a robust default without specialized implementations.

\section*{Funding}

This research was supported by the Israel Science Foundation (ISF), grant no. 353/24. The funding source had no role in the study design, analysis, interpretation of results, or decision to submit the article for publication.

\section*{Declaration of Competing Interest}

The author declares that he has no known competing financial interests or personal relationships that could have appeared to influence the work reported in this paper.

\paragraph{Declaration of generative AI and AI-assisted technologies in the manuscript preparation process:}

During the preparation of this work the author used ChatGPT and Claude for editing and analysis. After using this tool/service, the author reviewed and edited the content as needed and takes full responsibility for the content of the published article.

\bibliography{references}

\end{document}


\maketitle

\section{Full Reachability-Graph Incidence Matrix}
\subsection*{Overview}
This supplement provides the full node--arc incidence matrix $B\in\{-1,0,1\}^{24\times 50}$ for the reachability-graph example in the paper. Rows correspond to reachable markings (states) and columns correspond to directed edges (arrows). Each column has exactly a single $-1$ at its tail state and a single $+1$ at its head state.

\subsection*{Row Ordering (States)}
\begin{enumerate}[leftmargin=2.5em, itemsep=0.15em, topsep=0pt]
\item $[p_1,p'_1]$
\item $[p_1,p'_2]$
\item $[p_1,p'_3]$
\item $[p_1,p'_4]$
\item $[p_2,p_3,p'_1]$
\item $[p_2,p_3,p'_2]$
\item $[p_2,p_3,p'_3]$
\item $[p_2,p_3,p'_4]$
\item $[p_3,p_4,p'_1]$
\item $[p_3,p_4,p'_2]$
\item $[p_3,p_4,p'_3]$
\item $[p_3,p_4,p'_4]$
\item $[p_2,p_5,p'_1]$
\item $[p_2,p_5,p'_2]$
\item $[p_2,p_5,p'_3]$
\item $[p_2,p_5,p'_4]$
\item $[p_4,p_5,p'_1]$
\item $[p_4,p_5,p'_2]$
\item $[p_4,p_5,p'_3]$
\item $[p_4,p_5,p'_4]$
\item $[p_6,p'_1]$
\item $[p_6,p'_2]$
\item $[p_6,p'_3]$
\item $[p_6,p'_4]$
\end{enumerate}

\subsection*{Column Ordering (Edges)}
\begin{enumerate}[leftmargin=2.5em, itemsep=0.15em, topsep=0pt]
\item $e_{1}: \; [p_1,p'_1] \to [p_1,p'_2] \;\; (\gg,t'_1)$
\item $e_{2}: \; [p_1,p'_2] \to [p_1,p'_3] \;\; (\gg,t'_2)$
\item $e_{3}: \; [p_1,p'_3] \to [p_1,p'_4] \;\; (\gg,t'_3)$
\item $e_{4}: \; [p_2,p_3,p'_1] \to [p_2,p_3,p'_2] \;\; (\gg,t'_1)$
\item $e_{5}: \; [p_2,p_3,p'_2] \to [p_2,p_3,p'_3] \;\; (\gg,t'_2)$
\item $e_{6}: \; [p_2,p_3,p'_3] \to [p_2,p_3,p'_4] \;\; (\gg,t'_3)$
\item $e_{7}: \; [p_3,p_4,p'_1] \to [p_3,p_4,p'_2] \;\; (\gg,t'_1)$
\item $e_{8}: \; [p_3,p_4,p'_2] \to [p_3,p_4,p'_3] \;\; (\gg,t'_2)$
\item $e_{9}: \; [p_3,p_4,p'_3] \to [p_3,p_4,p'_4] \;\; (\gg,t'_3)$
\item $e_{10}: \; [p_2,p_5,p'_1] \to [p_2,p_5,p'_2] \;\; (\gg,t'_1)$
\item $e_{11}: \; [p_2,p_5,p'_2] \to [p_2,p_5,p'_3] \;\; (\gg,t'_2)$
\item $e_{12}: \; [p_2,p_5,p'_3] \to [p_2,p_5,p'_4] \;\; (\gg,t'_3)$
\item $e_{13}: \; [p_4,p_5,p'_1] \to [p_4,p_5,p'_2] \;\; (\gg,t'_1)$
\item $e_{14}: \; [p_4,p_5,p'_2] \to [p_4,p_5,p'_3] \;\; (\gg,t'_2)$
\item $e_{15}: \; [p_4,p_5,p'_3] \to [p_4,p_5,p'_4] \;\; (\gg,t'_3)$
\item $e_{16}: \; [p_6,p'_1] \to [p_6,p'_2] \;\; (\gg,t'_1)$
\item $e_{17}: \; [p_6,p'_2] \to [p_6,p'_3] \;\; (\gg,t'_2)$
\item $e_{18}: \; [p_6,p'_3] \to [p_6,p'_4] \;\; (\gg,t'_3)$
\item $e_{19}: \; [p_1,p'_1] \to [p_2,p_3,p'_1] \;\; (t_1,\gg)$
\item $e_{20}: \; [p_2,p_3,p'_1] \to [p_3,p_4,p'_1] \;\; (t_2,\gg)$
\item $e_{21}: \; [p_2,p_3,p'_1] \to [p_2,p_5,p'_1] \;\; (t_3,\gg)$
\item $e_{22}: \; [p_2,p_3,p'_1] \to [p_4,p_5,p'_1] \;\; (t_4,\gg)$
\item $e_{23}: \; [p_3,p_4,p'_1] \to [p_4,p_5,p'_1] \;\; (t_3,\gg)$
\item $e_{24}: \; [p_2,p_5,p'_1] \to [p_4,p_5,p'_1] \;\; (t_2,\gg)$
\item $e_{25}: \; [p_4,p_5,p'_1] \to [p_6,p'_1] \;\; (t_5,\gg)$
\item $e_{26}: \; [p_1,p'_2] \to [p_2,p_3,p'_2] \;\; (t_1,\gg)$
\item $e_{27}: \; [p_2,p_3,p'_2] \to [p_3,p_4,p'_2] \;\; (t_2,\gg)$
\item $e_{28}: \; [p_2,p_3,p'_2] \to [p_2,p_5,p'_2] \;\; (t_3,\gg)$
\item $e_{29}: \; [p_2,p_3,p'_2] \to [p_4,p_5,p'_2] \;\; (t_4,\gg)$
\item $e_{30}: \; [p_3,p_4,p'_2] \to [p_4,p_5,p'_2] \;\; (t_3,\gg)$
\item $e_{31}: \; [p_2,p_5,p'_2] \to [p_4,p_5,p'_2] \;\; (t_2,\gg)$
\item $e_{32}: \; [p_4,p_5,p'_2] \to [p_6,p'_2] \;\; (t_5,\gg)$
\item $e_{33}: \; [p_1,p'_3] \to [p_2,p_3,p'_3] \;\; (t_1,\gg)$
\item $e_{34}: \; [p_2,p_3,p'_3] \to [p_3,p_4,p'_3] \;\; (t_2,\gg)$
\item $e_{35}: \; [p_2,p_3,p'_3] \to [p_2,p_5,p'_3] \;\; (t_3,\gg)$
\item $e_{36}: \; [p_2,p_3,p'_3] \to [p_4,p_5,p'_3] \;\; (t_4,\gg)$
\item $e_{37}: \; [p_3,p_4,p'_3] \to [p_4,p_5,p'_3] \;\; (t_3,\gg)$
\item $e_{38}: \; [p_2,p_5,p'_3] \to [p_4,p_5,p'_3] \;\; (t_2,\gg)$
\item $e_{39}: \; [p_4,p_5,p'_3] \to [p_6,p'_3] \;\; (t_5,\gg)$
\item $e_{40}: \; [p_1,p'_4] \to [p_2,p_3,p'_4] \;\; (t_1,\gg)$
\item $e_{41}: \; [p_2,p_3,p'_4] \to [p_3,p_4,p'_4] \;\; (t_2,\gg)$
\item $e_{42}: \; [p_2,p_3,p'_4] \to [p_2,p_5,p'_4] \;\; (t_3,\gg)$
\item $e_{43}: \; [p_2,p_3,p'_4] \to [p_4,p_5,p'_4] \;\; (t_4,\gg)$
\item $e_{44}: \; [p_3,p_4,p'_4] \to [p_4,p_5,p'_4] \;\; (t_3,\gg)$
\item $e_{45}: \; [p_2,p_5,p'_4] \to [p_4,p_5,p'_4] \;\; (t_2,\gg)$
\item $e_{46}: \; [p_4,p_5,p'_4] \to [p_6,p'_4] \;\; (t_5,\gg)$
\item $e_{47}: \; [p_1,p'_1] \to [p_2,p_3,p'_2] \;\; (t_1,t'_1)$
\item $e_{48}: \; [p_2,p_3,p'_2] \to [p_3,p_4,p'_3] \;\; (t_2,t'_2)$
\item $e_{49}: \; [p_2,p_5,p'_2] \to [p_4,p_5,p'_3] \;\; (t_2,t'_2)$
\item $e_{50}: \; [p_4,p_5,p'_3] \to [p_6,p'_4] \;\; (t_5,t'_3)$
\end{enumerate}

\newpage

\subsection*{Incidence Matrix $B$ of the Reachability Graph}
\paragraph{Columns 1--25.}
\begin{center}
\renewcommand{\arraystretch}{1.0}
\resizebox{0.99\linewidth}{!}{%
\begin{tabular}{r|*{25}{r}}
 & 1&2&3&4&5&6&7&8&9&10&11&12&13&14&15&16&17&18&19&20&21&22&23&24&25\\
\hline
Row 1 & -1&0&0&0&0&0&0&0&0&0&0&0&0&0&0&0&0&0&-1&0&0&0&0&0&0\\
Row 2 & 1&-1&0&0&0&0&0&0&0&0&0&0&0&0&0&0&0&0&0&0&0&0&0&0&0\\
Row 3 & 0&1&-1&0&0&0&0&0&0&0&0&0&0&0&0&0&0&0&0&0&0&0&0&0&0\\
Row 4 & 0&0&1&0&0&0&0&0&0&0&0&0&0&0&0&0&0&0&0&0&0&0&0&0&0\\
Row 5 & 0&0&0&-1&0&0&0&0&0&0&0&0&0&0&0&0&0&0&1&-1&-1&-1&0&0&0\\
Row 6 & 0&0&0&1&-1&0&0&0&0&0&0&0&0&0&0&0&0&0&0&0&0&0&0&0&0\\
Row 7 & 0&0&0&0&1&-1&0&0&0&0&0&0&0&0&0&0&0&0&0&0&0&0&0&0&0\\
Row 8 & 0&0&0&0&0&1&0&0&0&0&0&0&0&0&0&0&0&0&0&0&0&0&0&0&0\\
Row 9 & 0&0&0&0&0&0&-1&0&0&0&0&0&0&0&0&0&0&0&0&1&0&0&-1&0&0\\
Row 10 & 0&0&0&0&0&0&1&-1&0&0&0&0&0&0&0&0&0&0&0&0&0&0&0&0&0\\
Row 11 & 0&0&0&0&0&0&0&1&-1&0&0&0&0&0&0&0&0&0&0&0&0&0&0&0&0\\
Row 12 & 0&0&0&0&0&0&0&0&1&0&0&0&0&0&0&0&0&0&0&0&0&0&0&0&0\\
Row 13 & 0&0&0&0&0&0&0&0&0&-1&0&0&0&0&0&0&0&0&0&0&1&0&0&-1&0\\
Row 14 & 0&0&0&0&0&0&0&0&0&1&-1&0&0&0&0&0&0&0&0&0&0&0&0&0&0\\
Row 15 & 0&0&0&0&0&0&0&0&0&0&1&-1&0&0&0&0&0&0&0&0&0&0&0&0&0\\
Row 16 & 0&0&0&0&0&0&0&0&0&0&0&1&0&0&0&0&0&0&0&0&0&0&0&0&0\\
Row 17 & 0&0&0&0&0&0&0&0&0&0&0&0&-1&0&0&0&0&0&0&0&0&1&1&1&-1\\
Row 18 & 0&0&0&0&0&0&0&0&0&0&0&0&1&-1&0&0&0&0&0&0&0&0&0&0&0\\
Row 19 & 0&0&0&0&0&0&0&0&0&0&0&0&0&1&-1&0&0&0&0&0&0&0&0&0&0\\
Row 20 & 0&0&0&0&0&0&0&0&0&0&0&0&0&0&1&0&0&0&0&0&0&0&0&0&0\\
Row 21 & 0&0&0&0&0&0&0&0&0&0&0&0&0&0&0&-1&0&0&0&0&0&0&0&0&1\\
Row 22 & 0&0&0&0&0&0&0&0&0&0&0&0&0&0&0&1&-1&0&0&0&0&0&0&0&0\\
Row 23 & 0&0&0&0&0&0&0&0&0&0&0&0&0&0&0&0&1&-1&0&0&0&0&0&0&0\\
Row 24 & 0&0&0&0&0&0&0&0&0&0&0&0&0&0&0&0&0&1&0&0&0&0&0&0&0\\
\end{tabular}}
\end{center}

\paragraph{Columns 26--50.}
\begin{center}
\renewcommand{\arraystretch}{1.0}
\resizebox{0.99\linewidth}{!}{%
\begin{tabular}{r|*{25}{r}}
 & 26&27&28&29&30&31&32&33&34&35&36&37&38&39&40&41&42&43&44&45&46&47&48&49&50\\
\hline
Row 1 & 0&0&0&0&0&0&0&0&0&0&0&0&0&0&0&0&0&0&0&0&0&-1&0&0&0\\
Row 2 & -1&0&0&0&0&0&0&0&0&0&0&0&0&0&0&0&0&0&0&0&0&0&0&0&0\\
Row 3 & 0&0&0&0&0&0&0&-1&0&0&0&0&0&0&0&0&0&0&0&0&0&0&0&0&0\\
Row 4 & 0&0&0&0&0&0&0&0&0&0&0&0&0&0&-1&0&0&0&0&0&0&0&0&0&0\\
Row 5 & 0&0&0&0&0&0&0&0&0&0&0&0&0&0&0&0&0&0&0&0&0&0&0&0&0\\
Row 6 & 1&-1&-1&-1&0&0&0&0&0&0&0&0&0&0&0&0&0&0&0&0&0&1&-1&0&0\\
Row 7 & 0&0&0&0&0&0&0&1&-1&-1&-1&0&0&0&0&0&0&0&0&0&0&0&0&0&0\\
Row 8 & 0&0&0&0&0&0&0&0&0&0&0&0&0&0&1&-1&-1&-1&0&0&0&0&0&0&0\\
Row 9 & 0&0&0&0&0&0&0&0&0&0&0&0&0&0&0&0&0&0&0&0&0&0&0&0&0\\
Row 10 & 0&1&0&0&-1&0&0&0&0&0&0&0&0&0&0&0&0&0&0&0&0&0&0&0&0\\
Row 11 & 0&0&0&0&0&0&0&0&1&0&0&-1&0&0&0&0&0&0&0&0&0&0&1&0&0\\
Row 12 & 0&0&0&0&0&0&0&0&0&0&0&0&0&0&0&1&0&0&-1&0&0&0&0&0&0\\
Row 13 & 0&0&0&0&0&0&0&0&0&0&0&0&0&0&0&0&0&0&0&0&0&0&0&0&0\\
Row 14 & 0&0&1&0&0&-1&0&0&0&0&0&0&0&0&0&0&0&0&0&0&0&0&0&-1&0\\
Row 15 & 0&0&0&0&0&0&0&0&0&1&0&0&-1&0&0&0&0&0&0&0&0&0&0&0&0\\
Row 16 & 0&0&0&0&0&0&0&0&0&0&0&0&0&0&0&0&1&0&0&-1&0&0&0&0&0\\
Row 17 & 0&0&0&0&0&0&0&0&0&0&0&0&0&0&0&0&0&0&0&0&0&0&0&0&0\\
Row 18 & 0&0&0&1&1&1&-1&0&0&0&0&0&0&0&0&0&0&0&0&0&0&0&0&0&0\\
Row 19 & 0&0&0&0&0&0&0&0&0&0&1&1&1&-1&0&0&0&0&0&0&0&0&0&1&-1\\
Row 20 & 0&0&0&0&0&0&0&0&0&0&0&0&0&0&0&0&0&1&1&1&-1&0&0&0&0\\
Row 21 & 0&0&0&0&0&0&0&0&0&0&0&0&0&0&0&0&0&0&0&0&0&0&0&0&0\\
Row 22 & 0&0&0&0&0&0&1&0&0&0&0&0&0&0&0&0&0&0&0&0&0&0&0&0&0\\
Row 23 & 0&0&0&0&0&0&0&0&0&0&0&0&0&1&0&0&0&0&0&0&0&0&0&0&0\\
Row 24 & 0&0&0&0&0&0&0&0&0&0&0&0&0&0&0&0&0&0&0&0&1&0&0&0&1\\
\end{tabular}}
\end{center}